\documentclass[10pt,twocolumn,letterpaper]{article}

\usepackage[pagenumbers]{cvpr} %
\usepackage[dvipsnames]{xcolor}

\newif\ifdraft

\draftfalse

\ifdraft

\newcommand{\asc}[1]{{\color{blue}[\textbf{Arik:} #1]}}
\newcommand{\dcc}[1]{{\color{red}[\textbf{Danny:} #1]}}
\newcommand{\moab}[1]{{\color{purple}[\textbf{Moab:} #1]}}
\newcommand{\avc}[1]{{\color{green}[\textbf{Andrey} #1]}}
\newcommand{\ahc}[1]{{\color{orange}[\textbf{Amir:} #1]}}
\newcommand{\sfc}[1]{{\color{red}[\textbf{Shlomi:} #1]}}

\newcommand{\sf}[1]{{\color{red}#1}}

\else

\newcommand{\asc}[1]{}
\newcommand{\dcc}[1]{}
\newcommand{\moab}[1]{}
\newcommand{\avc}[1]{}
\newcommand{\ahc}[1]{}
\newcommand{\sfc}[1]{}

\fi

\definecolor{cvprblue}{rgb}{0.21,0.49,0.74}
\usepackage[pagebackref,breaklinks,colorlinks,citecolor=cvprblue]{hyperref}
\newcommand*{\affaddr}[1]{#1} 
\newcommand*{\affmark}[1][*]{\textsuperscript{#1}}

\usepackage[symbol]{footmisc}
\usepackage{makecell}
\def\paperID{4534} %
\def\confName{CVPR}
\def\confYear{2024}

\title{
PALP: Prompt Aligned Personalization of Text-to-Image Models
}
\author{%
\large
 Moab Arar\footnote[1]~~\affmark[1,2], Andrey Voynov\affmark[2], Amir Hertz\affmark[2], Omri Avrahami\footnote[1]~~\affmark[2,4],\\  Shlomi Fruchter\affmark[1], Yael Pritch~\affmark[2], Daniel Cohen-Or\footnote[1] ~~\affmark[1,2], Ariel Shamir\footnote[1]~~\affmark[2,3]\\
\normalsize{\affaddr{\affmark[1] Tel-Aviv University}},  \normalsize{\affaddr{\affmark[2] Google Research}} 
\normalsize{\affaddr{\affmark[3] Reichman University}}, \normalsize{\affaddr{\affmark[4] The Hebrew University of Jerusalem}}\\
}

\begin{document}
\twocolumn[{%
\renewcommand\twocolumn[1][]{#1}%
\maketitle
\begin{center}
    \centering
    \captionsetup{type=figure}
    \includegraphics[width=.99\textwidth]{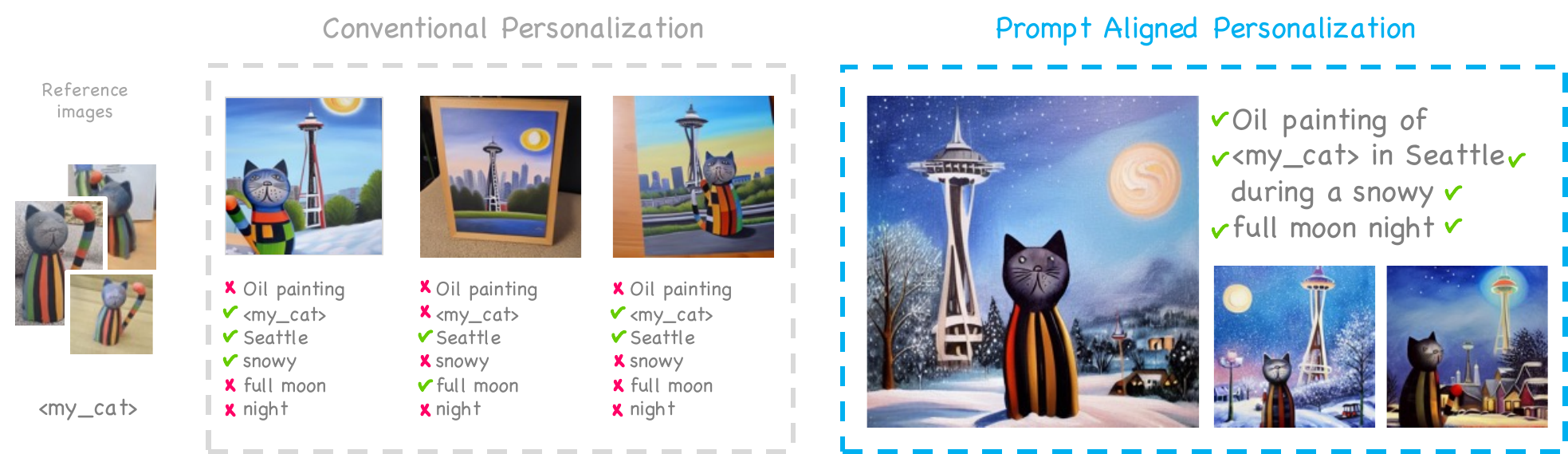}
    \captionof{figure}{Prompt aligned personalization allow rich and complex scene generation, including all elements of a condition prompt (right).}
    \label{fig:teaser}
\end{center}%
}]

\begin{abstract}
Content creators often aim to create personalized images using personal subjects that go beyond the capabilities of conventional text-to-image models. Additionally, they may want the resulting image to encompass a specific location, style, ambiance, and more. Existing personalization methods may compromise personalization ability or the alignment to complex textual prompts. This trade-off can impede the fulfillment of user prompts and subject fidelity. We propose a new approach focusing on personalization methods for a \emph{single} prompt to address this issue. We term our approach prompt-aligned personalization. While this may seem restrictive, our method excels in improving text alignment, enabling the creation of images with complex and intricate prompts, which may pose a challenge for current techniques. In particular, our method keeps the personalized model aligned with a target prompt using an additional score distillation sampling term. We demonstrate the versatility of our method in multi- and single-shot settings and further show that it can compose multiple subjects or use inspiration from reference images, such as artworks. We compare our approach quantitatively and qualitatively with existing baselines and state-of-the-art techniques.
\end{abstract}

\footnotetext[1]{Work done while working at Google.}
\footnotetext[2]{Project page available at \url{https://prompt-aligned.github.io/.}}

\section{Introduction}

Text-to-image models have shown exceptional abilities to generate a diversity of images in various settings (place, time, style, and appearances), such as "a sketch of Paris on a rainy day" or "a Manga drawing of a teddy bear at night"~\cite{DALLE2, IMAGEN}. Recently, personalization methods even allow one to include specific subjects (objects, animals, or people) into the generated images~\cite{TI, DB}. In practice, however, such models are difficult to control and may require significant prompt engineering and re-sampling to create the specific image one has in mind. It is even more acute with personalized models, where it is challenging to include the personal item or character in the image and simultaneously fulfill the textual prompt describing the content and style. This work proposes a method for better personalization and prompt alignment, especially suited for complex prompts.

\begin{figure}
    \centering
    \includegraphics[width=\linewidth]{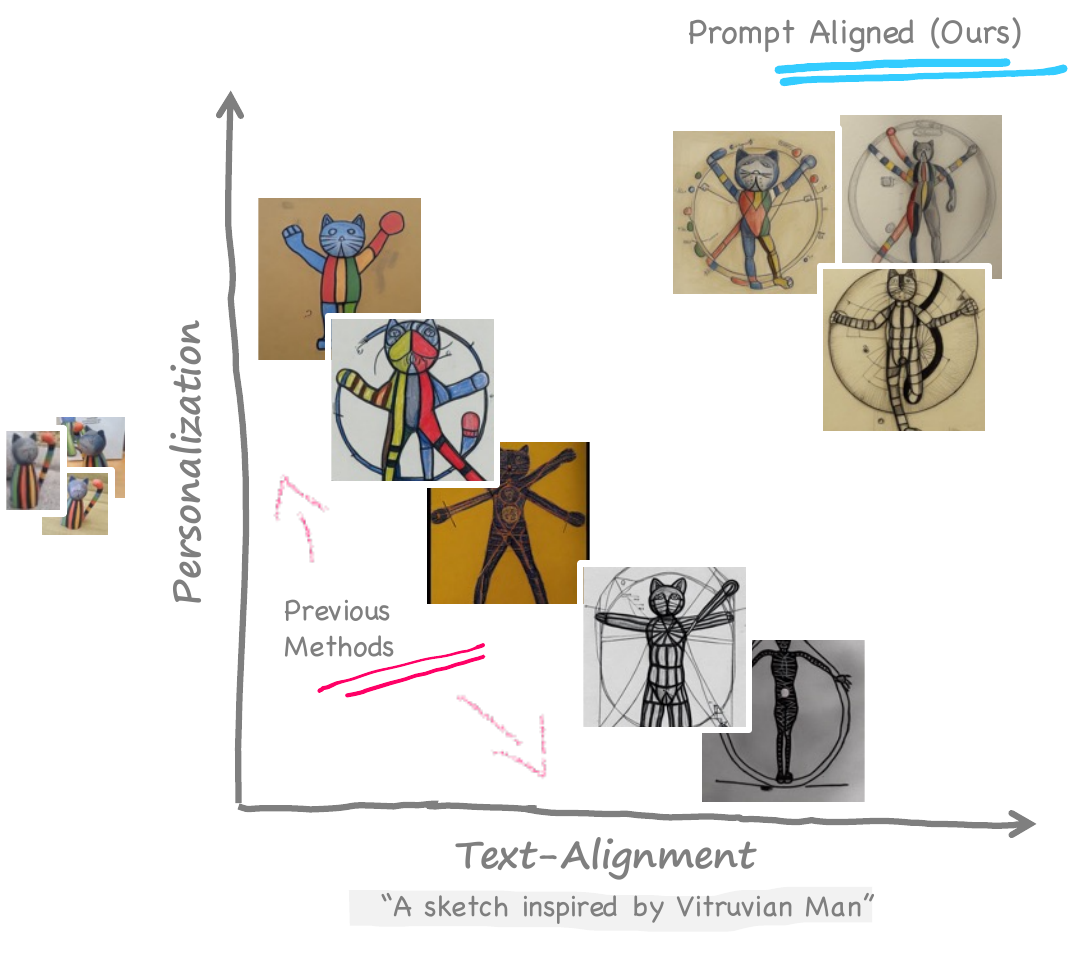}
    \caption{Previous personalization methods struggle with complex prompts (e.g., ``A sketch inspired by Vitruvian man'') presenting a trade-off between prompt-alignment and subject-fidelity. Our method, optimizes for both, without compromising either.}
    \label{fig:us_vs_previous_on_davnici}
\end{figure}

A key ingredient in personalization methods is fine-tuning pre-trained text-to-image models on a small set of personal images while relying on heavy regularization to maintain the model's capacity. Doing so will preserve the model's prior knowledge and allow the user to synthesize images with various prompts; however, it impairs capturing identifying features of the target subject. On the other hand, persisting on identification accuracy can hinder the prompt-alignment capabilities. Generally speaking, the trade-off between identity preservation and prompt alignment is a core challenge in personalization methods (see~\cref{fig:us_vs_previous_on_davnici}).

Content creators and AI artists frequently have a clear idea of the prompt they wish to utilize. It may involve stylization and other factors that current personalization methods struggle to maintain. Therefore, we take a different approach by focusing on excelling with a \textit{single} prompt rather than offering a general-purpose method intended to perform well with a wide range of prompts. This approach enables both (i) learning the unique features of the subject from a few or even a single input image and (ii) generating richer scenes that are better aligned with the user's desired prompt (see~\cref{fig:teaser}).

Our work is based on the premise that existing models possess knowledge of all elements within the target prompt except the new personal subject. Consequently, we leverage the pre-trained model's prior knowledge to prevent personalized models from losing their understanding of the target prompt. In particular, since we know the target prompt during training, we show how to incorporate score-distillation guidance~\cite{SDS} to constrain the personalized model's prediction to stay aligned with the pre-trained one. Therefore, we introduce a framework comprising two components: personalization, which teaches the model about our new subject, and prompt alignment which prevents it from forgetting elements included in the target prompt.

\begin{figure}
    \centering
    \includegraphics[width=\linewidth]{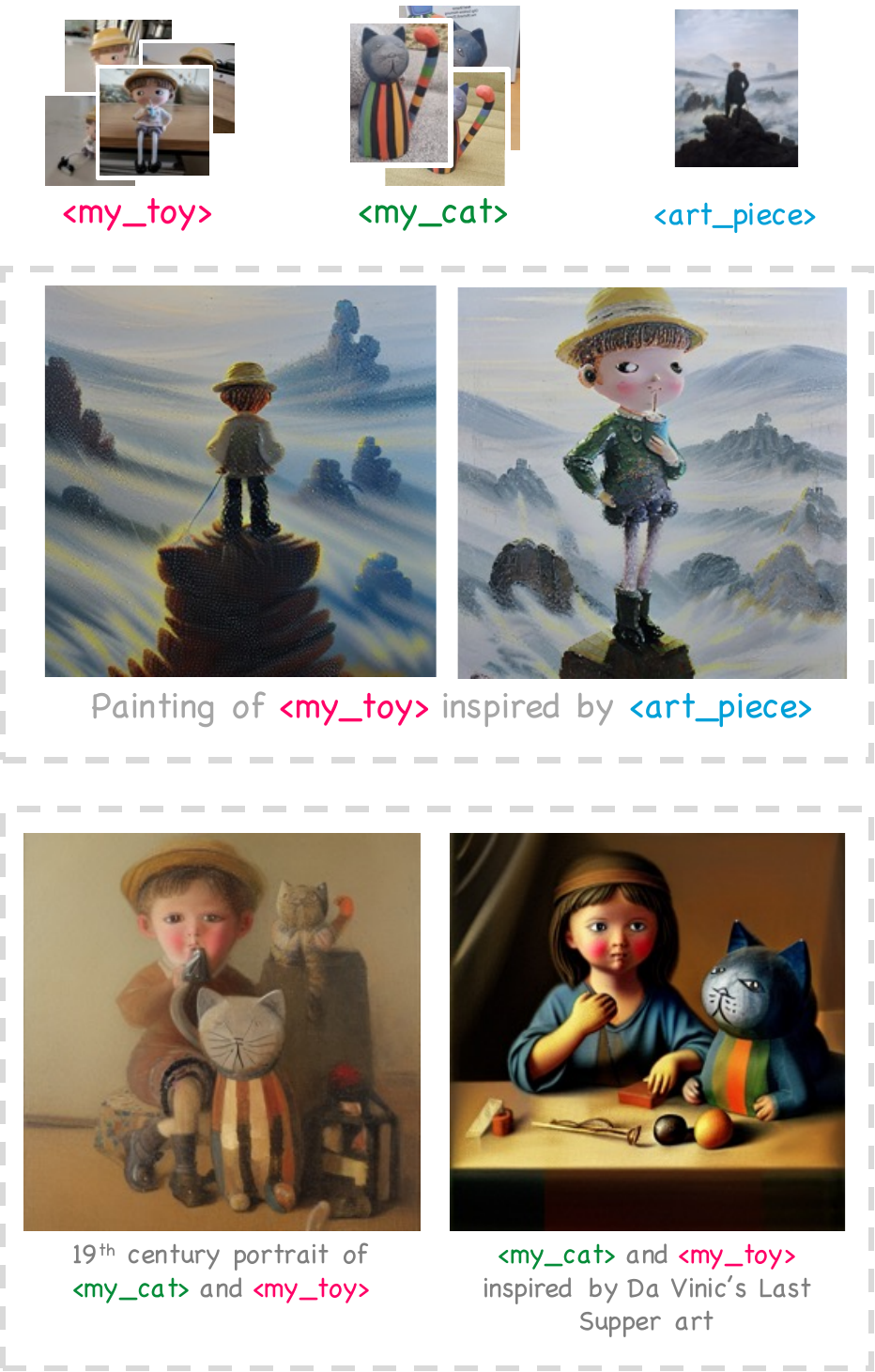}
    \caption{PALP for multi-subject personalization achieves coherent and prompt-aligned results. Our method works when the subject has only one image (e.g., the "Wanderer above the Sea of Fog" artwork by Caspar David Friedrich).}
    \label{fig:composition}
\end{figure}

Our approach liberates content creators from constraints associated with specific prompts, unleashing the full potential of text-to-image models. We evaluate our method qualitatively and quantitatively. We show superior results compared with the baselines in multi- and single-shot settings, all without pre-training on large-scale data~\cite{DATE, E4T}, which can be difficult for certain domains. Finally, we show that our method can accommodate multi-subject personalization with minor modification and offer new applications such as drawing inspiration from a single artistic painting, and not just text (see Figure~\ref{fig:composition}).

\section{Related work}

\paragraph{Text-to-image synthesis} has marked an unprecedented progress in recent years~\cite{DALLE2, LDM, GLIDE, IMAGEN, MAKE_A_SCENE}, mostly due to large-scale training on data like LAION-400m~\cite{LAION}. Our approach uses pre-trained diffusion models~\cite{DDPM} to extend their understanding to new subjects. We use the publicly available Stable-Diffusion model~\cite{LDM} for most of our experiments since baseline models are mostly open-source on SD. We further verify our method on a larger latent diffusion model variant~\cite{LDM}.

\paragraph{Text-based editing} methods rely on contrastive multi-modal models like CLIP~\cite{CLIP} as an interface to guide local and global edits~\cite{StyleCLIP, BlendedLatentDiffusion, BlendedDiffusion, StyleGAN_NADA, Text2Live}. Recently, Prompt-to-Prompt (P2P)~\cite{Prompt2Prompt} was proposed as a way to edit and manipulate \emph{generated} images by editing the attention maps in the cross-attention layers of a pre-trained text-to-image model. Later,~\citet{NullInversion} extended P2P for real images by encoding them into the null-conditioning space of classifier-free guidance~\cite{CFG}. InstructPix2Pix~\cite{InstructPix2Pix} uses an instruction-guided image-to-image translation network trained on synthetic data. Others preserve image-structure by using reference attention maps~\cite{ZeroShotI2I} or features extracted using DDIM~\cite{DDIM} inversion~\cite{PlugAndPlay}. Imagic~\cite{Imagic} starts from a target prompt, finds text embedding to reconstruct an input image, and later interpolates between the two to achieve the final edit. UniTune~\cite{UniTune}, on the other hand, performs the interpolation in pixel space during the denoising backward process. In our work, we focus on the ability to generate images depicting a given subject, which may not necessarily maintain the global structure of an input image.

\paragraph{Early personalization methods} like Textual Inversion~\cite{TI} and DreamBooth~\cite{DB} tune pre-trained text-2-image models to represent new subjects, either by finding a new soft word-embedding~\cite{TI} or calibrating model-weights~\cite{DB} with existing words to represent the newly added subject. Later methods improved memory requirements of previous methods using Low-Rank updates~\cite{LORA, CustomDiffusion, KeyLocked, simoLoRA2023} or compact-parameter space~\cite{SVDiff}. In another axis, NeTI~\cite{NeTI} and $P+$~\cite{PPLUS} extend TI~\cite{TI} using more tokens to capture the subject-identifying features better. Personalization can also be used for other tasks. ReVersion~\cite{ReVersion} showed how to learn relational features from reference images, and~\citet{ConceptDecompose} used personalization to decompose and visualize concepts at different abstraction levels. Chefer et. al~\cite{chefer2023hidden} propose an interpretability method for text-to-image models by decomposing concepts into interpretable tokens. Another line-of-works pre-train encoders on large-data for near-instant, single-shot adaptation~\cite{E4T, DATE, IPAdapter, Face0, SuTI, ProFusion, ELITE}. Single-image personalization has also been addressed in~\citet{BreakAScene}, where the authors use segmentation masks to personalize a model on different subjects. In our work, we focus on prompt alignment, and any of the previous personalization methods may be replaced by our baseline personalization method.

\paragraph{Score Distillation Sampling (SDS)} emerged as a technique for leveraging 2D-diffusion models priors~\cite{LDM, IMAGEN} for 3D-generation from textual input. Soon, this technique found way to different applications like SVG generation~\cite{VectorFusion, WordAsImage}, image-editing~\cite{DDS}, and more~\cite{DomainAdaptationGANSDS}. Other variant of SDS~\cite{SDS} aim to improve the image-generation quality of SDS, which suffers from over-saturation and blurriness~\cite{NFSD, ProlificDreamer}. In our approach, we propose a framework that leverages score sampling to maintain alignment with the target prompt. Alternative score-sampling techniques can be considered to further boost the text-alignment of the personalized method.

\paragraph{Text-to-image alignment} methods address text-related issues that arise in base diffusion generative models. These issues include neglecting specific text parts, attribute mixing, and more. Previous methods address these issues through attention-map re-weighting\cite{StructureDiffusion, HarnessingForTextAlignment, GroundedTextAlignment}, latent-optimization~\cite{AttendAndExcite, LinguisticBinding}, or re-training with additional data~\cite{RECAP}. However, none of these methods address prompt alignment of personalization methods, instead, they aim to enhance the base models in generating text-aligned images.

\section{Preliminaries}
\label{sec:personalization}
\begin{figure*}
    \centering
    \includegraphics[width=0.95\linewidth]{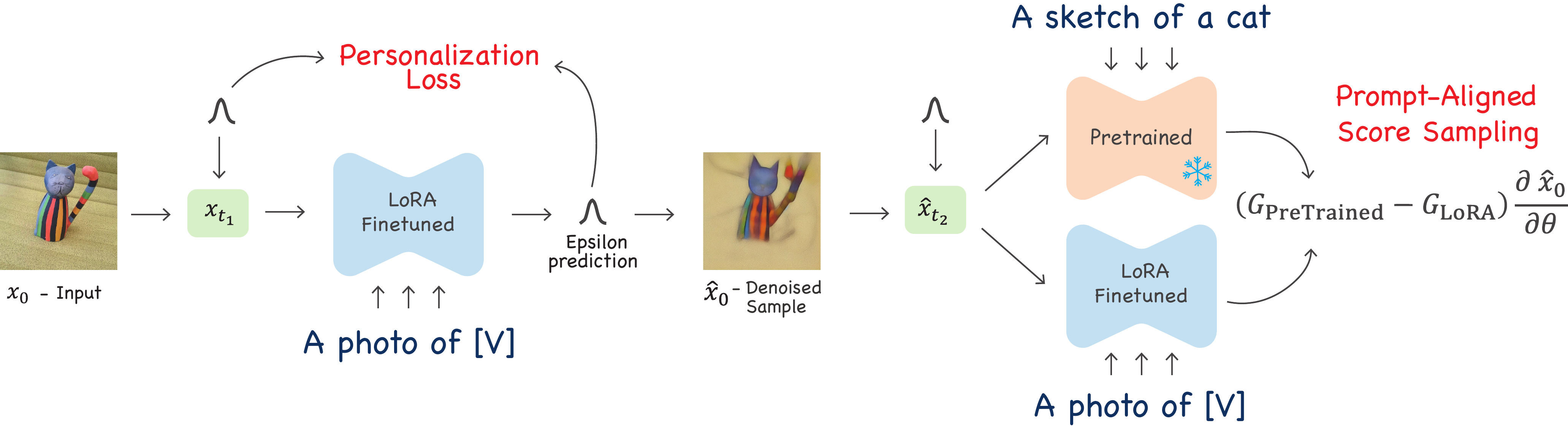}
    \caption{Method overview. We propose a framework consisting of a personalization path (left) and a prompt-alignment branch (right) applied simultaneously in the same training step. We achieve personalization by finetuning the pre-trained model using a simple reconstruction loss to denoise the new subject $S$. To keep the model aligned with the target prompt, we additionally use score sampling to pivot the prediction towards the direction of the target prompt $y$, e.g., "A sketch of a cat." In this example, when personalization and text alignment are optimized simultaneously, the network learns to denoise the subject towards a "sketch" like representation. Finally, our method does not induce a significant memory overhead due to the efficient estimation of the score function, following~\cite{SDS}.}
    \label{fig:overview}
\end{figure*}

\paragraph{Generative Diffusion Models.} 
Diffusion models perform a backward diffusion process to generate an image. In this process, the diffusion model $G$ progressively denoises a source $x_T \sim \mathcal{N}(0,1)$ to produce a real sample $x_0$ from an underlying data distribution $p(X)$. At each timestep $t \in \{0,1,...,T\}$, the model predicts a noise $\hat{\epsilon} = G\left(x_t, t, y; \theta\right)$ conditioned on a prompt $y$ and timestep $t$. The generative model is trained by maximizing the evidence lower bound (ELBO) using a denoising score matching objective~\cite{DDPM}:

\begin{equation}
\label{eq:diff}
\mathcal{L}(\textbf{x},y) = \mathbb{E}_{t \sim [0,T], \epsilon \sim \mathcal{N}(0,1)} \left[\lVert G\left(x_t, t, y; \theta\right) - \epsilon \rVert_{2}^{2} \right].
\end{equation}

Here, $x_t = \sqrt{\bar{\alpha}_{t}} \textbf{x} + \sqrt{1-\bar{\alpha}_{t}}\epsilon$, and $\textbf{x}$ is a sample from the real-data distribution $p(X)$. Throughout this section, we will write $G_{\theta}(x_t, y) = G\left(x_t, t, y; \theta\right)$.

\paragraph{Personalization.} Personalization methods fine-tune the diffusion model $G$ for a new target subject $S$, by optimizing~\cref{eq:diff} on a small set of images representing $S$. Textual-Inversion~\cite{TI} optimizes new word embeddings to represent the new subject. This is done by pairing a generic prompt with the placeholder $[V]$, e.g., ``A photo of [V]'', where $[V]$ is mapped to the newly-added word embedding. DreamBooth~\cite{DB} calibrates existing word-embeddings to represent the personal subject $S$. This can be done by adjusting the model weights, or using recent more efficient methods that employ a Low-Rank Adaptation (LoRA)~\cite{LORA} on a smaller subset of weights~\cite{CustomDiffusion, simoLoRA2023, KeyLocked}.

\section{Prompt Alignment Method}
\label{sec:method}

\subsection{Overview}
Our primary objective is to teach $G$ to generate images related to a new subject $S$. However, unlike previous methods, we strongly emphasize achieving optimal results for a \emph{single textual prompt}, denoted by $y$. For instance, consider a scenario where the subject of interest is one's cat, and the prompt is $y = \text{``A sketch of [my cat] in Paris.''}$ In this case, we aim to generate an image that faithfully represents the cat while incorporating all prompt elements, including the sketchiness and the Parisian context. 

The key idea is to optimize two objectives: personalization and prompt alignment. For the first part, i.e., personalization, we fine-tune $G$ to reconstruct $S$ from noisy samples. One option for improving prompt alignment would be using our target prompt $y$ as a condition when tuning $G$ on $S$. However, this results in sub-optimal outcome since we have no images depicting $y$ (e.g., we have no sketch images of our cat, nor photos of it in Paris). Instead, we push $G$'s noise prediction towards the target prompt $y$. In particular, we steer $G$'s estimation of $S$ towards the distribution density $p(x|y)$ using score-sampling methods (See~\cref{fig:overview}). By employing both objectives simultaneously, we achieve two things: (1) the ability to generate the target subject $S$ via the backward-diffusion process (personalization) while (2) ensuring the noise-predictions are aligned with text $y$. We next explain our method in detail.

\subsection{Personalization}
We follow previous methods~\cite{TI, DB, CustomDiffusion, KeyLocked}, and fine-tune $G$ on a small set of images representing $S$, which can be as small as a single photo. We update $G$'s weights using LoRA method~\cite{LORA}, where we only update a subset of the network's weights $\theta_{\text{LoRA}} \subseteq \theta $, namely, the self- and cross- attention layers. Furthermore, we optimize a new word embedding $[V]$ to represent $S$.

\subsection{Prompt-Aligned Score Sampling}

Long fine-tuning on a small image set could cause the model to overfit. In this case, the diffusion model predictions always steer the backward denoising process towards one of the training images, regardless of any conditional prompt. Indeed, we observed that overfitted models could estimate the inputs from noisy samples, even in a single denoising step. 

\begin{figure}

    \centering
    \begin{subfigure}[t]{0.15\linewidth}
    \includegraphics[width=\textwidth]{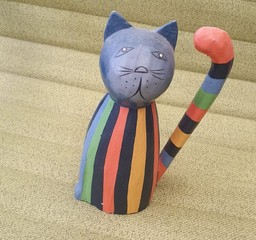}
    \caption{Input}
    \end{subfigure}
    \centering
    \begin{subfigure}[t]{0.26\linewidth}
    \includegraphics[width=\textwidth]{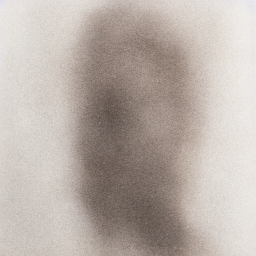}
    \caption{base model}
    \end{subfigure}
    \centering
    \begin{subfigure}[t]{0.26\linewidth}
    \includegraphics[width=\textwidth]{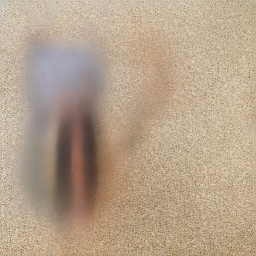}
    \caption{Over-fitted}
    \end{subfigure}
    \centering
    \begin{subfigure}[t]{0.26\linewidth}
    \includegraphics[width=\textwidth]{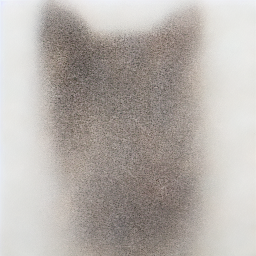}
    \caption{Prompt aligned}
    \end{subfigure}
    
    \caption{Visualization of $\hat{x}_0$. We visualize the model estimation of $\hat{x}_0$ given a pure-noise and the prompt "A sketch of [V]." The base model (b) is not personalized to the target subject and predicts mainly the "Sketch" appearance. Personalization methods (c) tend to overfit the input image where many image elements, including the background and the subject colors are restored, suggesting the model does not consider the prompt condition. Prompt aligned personalization (c) maintains the sketchiness and does not overfit (see cat-like shape).}
    \label{fig:methodvisx0hat}
\end{figure}

In~\cref{fig:methodvisx0hat}, we visualize the model prediction by analyzing its estimate of $x_0$, the real-sample image. The real-sample estimation $\hat{x}_0$ is derived from the model's noise prediction via:
\begin{equation}
\label{eq:x0_hat}
   \hat{x}_0  = \frac{x_t - \sqrt{1-\bar{\alpha}_{t}}G_{\theta}(x_t,y) }{ \sqrt{\bar{\alpha}_{t}} }.
\end{equation}.

As can be seen from~\cref{fig:methodvisx0hat}, the pre-trained model (before personalization) steers the image towards a sketch-like image, where the estimate $\hat{x}_0$ has a white background, but the subject details are missing. Apply personalization without PALP overfits, where elements from the training images, like the background, are more dominant, and sketchiness fades away, suggesting miss-alignment with the prompt ``A sketch'' and over-fitting to the input image. Using PALP, the model prediction steers the backward-denoising process towards a sketch-like image, while staying personalized, where a cat like shape is restored.

Our key idea is to encourage the model's denoising prediction towards the target prompt. Alternatively, we push the model's estimation of the real-sample, denoted by $\hat{x}_0$, to be aligned with the prompt $y$. 

In our example, where the target prompt is "A sketch of [V]," we push the model prediction toward a sketch and prevent overfitting a specific image background. Together with the personalization loss, this will encourage the model to focus on capturing [V]'s identifying features rather than reconstructing the background and other distracting features. 

To achieve the latter, we use Score Distillation Sampling (SDS) techniques~\cite{SDS}. In particular, given a clean target prompt $y^c$, that doesn't contain the placeholder "[V]". Then we use score sampling to optimize the loss:
\begin{equation}
\label{eq:palp_loss}
\mathcal{L}(\hat{x}_0,y^c),
\end{equation}
from~\cref{eq:diff}. Note, we use the pre-trained network weights to evaluate~\cref{eq:palp_loss}, and omit all learned placeholder (e.g., $[V]$). Therefore, this ensures that by minimizing the loss, we will stay aligned with the textual prompt $y$ since the pre-trained model possesses all knowledge about the prompt elements.

Poole et al.~\cite{SDS} found an elegant and efficient approximation of the score function using:
\begin{equation}
\label{eq:sds}
\nabla \mathcal{L}_{SDS}(\textbf{x}) =  \Tilde{w}(t) \left(G^{\alpha}\left(x_t, t, y^c; \theta\right) - \epsilon\right) \frac{\partial \textbf{x}}{\partial \phi},
\end{equation}
where $\phi$ are the weights controlling the appearance of $\textbf{x}$, and $\Tilde{w}(t)$ is a weighting function. Here $G^{\alpha}$ denotes the classifier-free guidance prediction, which is an extrapolation of the conditional and unconditioned ($y=\emptyset$) noise prediction. The scalar $\alpha \in \mathbb{R}^+$ controls the extrapolation via:
\begin{equation}
\label{eq:cfg}
    G^{\alpha}\left(x_t, t, y^c; \theta\right) =  (1-\alpha) \cdot G_{\theta}\left(x_t, \emptyset \right) + \alpha \cdot G_{\theta}\left(x_t, y^c \right).
\end{equation}

In our case, $\textbf{x} = \hat{x}_{0}$, and it is derived from $G$'s noise prediction as per~\cref{eq:x0_hat} . Therefore, the appearance of $\hat{x}_0$ is directly controlled by the LoRA weights and $[V]$.

\subsection{Avoiding Over-saturation and Mode Collapse}
\label{sec:PALP_score}
Guidance by SDS produces less diverse and over-saturated results. Alternative implementations like~\cite{ProlificDreamer, NFSD} improved diversity, but the overall personalization is still affected (see ~\cref{sec:a_sds_vs_ours}). We argue that the reason behind this is that the loss in~\cref{eq:sds} pushes the prediction $\hat{x}_0$ towards the center of distribution of $p(x|y^c)$, and since the clean target prompt $y^c$ contains only the subject class, i.e., ``A sketch of a cat'', then as a result, the loss will encourage $\hat{x}_0$ towards the center of distribution of a general cat, not our one. 

Instead, we found the Delta Denoising Score (DDS)~\cite{DDS} variant to work better. In particular, we use the residual direction between the personalized model's prediction $G^\beta_{\theta_{LoRA}} (\hat{x}_t, y_{P})$, and the pre-trained one $G^\alpha_{\theta}(\hat{x}_t, y^c)$, where $\alpha$ and $\beta$ are their guidance scales, respectively. Here $y_P$ is the personalization prompt, i.e., "A photo of [V]," and $y_c$ is a clean prompt, e.g., "A sketch of a cat."  The score can then be estimated by: 
\begin{equation}
\label{eq:palp_grad}
\nabla \mathcal{L}_{PALP} = \Tilde{w}(t) ( G^\alpha_{\theta}(\hat{x}_t, y^c) - G^\beta_{\theta_{LoRA}} (\hat{x}_t, y_{P})) \frac{\partial \hat{x}_0}{\partial \theta_{\text{LoRA}}},
\end{equation} which perturbs the denoising prediction towards the target prompt (see right part of~\cref{fig:overview}). Our experiments found imbalanced guidance scales, i.e., $\alpha > \beta$, to perform better. We have also considered two variant implementations: (1) in the first, we use the same noise for the two branches, i.e., the text-alignment and personalization branches, and (2) we use two i.i.d noise samples for the branches. Using the same noise achieves better text alignment than the latter variant.

\paragraph{On the Computational Complexity of PALP: } the gradient in~\cref{eq:palp_grad}, is proportional to the personalization gradient with the scaling:
\begin{equation}
     \frac{\partial \hat{x}_0}{\partial \theta_{\text{LoRA}}} \propto  - \frac{\sqrt{1-\bar{\alpha}_{t}}}{ \sqrt{\bar{\alpha}_{t}} } \nabla G_{\theta_{LoRA}}(x_t,y_P).
\end{equation} 
and since the gradient $\nabla G_{\theta_{LoRA}}(x_t,y_P)$ is also calculated as part of the derivative of~\cref{eq:diff}, then, we do not need to back-propagate through the text-to-image model through the prompt-alignment branch. This may be useful to use guidance from bigger models to boost smaller model's personalization performance. Finally, we also note that the scaling term is very high for large $t$-values. Re-scaling the gradient back by $  \sqrt{\bar{\alpha}_{t}} / \sqrt{1-\bar{\alpha}_{t}} $ ensures uniform gradient update across all timesteps and improves numerical stability.

\section{Results}

\begingroup

\setlength{\tabcolsep}{2pt} %
\renewcommand{\arraystretch}{1.0} %

\begin{table*}
        \centering
        \begin{tabular}{cccccccc}
             Input &  \multicolumn{2}{c}{Ours} &  \hspace{2pt} &  TI+DB~\cite{TI, DB} &  CD~\cite{CustomDiffusion} & $P+$~\cite{PPLUS} & NeTI~\cite{NeTI} \\[5 pt]
             \includegraphics[height=2.3cm, width=2.3cm]{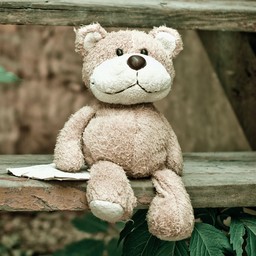} &  \includegraphics[height=2.3cm, width=2.3cm]{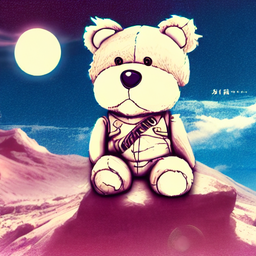} &  \includegraphics[height=2.3cm, width=2.3cm]{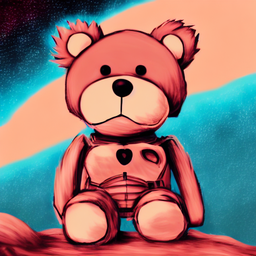} &   &  \includegraphics[height=2.3cm, width=2.3cm]{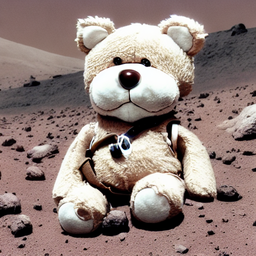} &  \includegraphics[height=2.3cm, width=2.3cm]{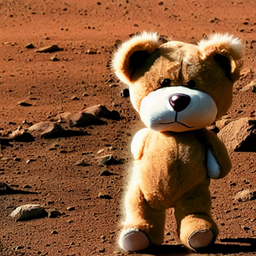} &  \includegraphics[height=2.3cm, width=2.3cm]{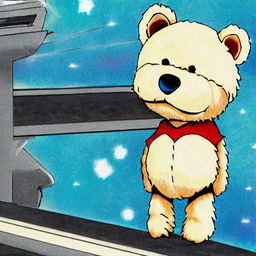} & \includegraphics[height=2.3cm, width=2.3cm]{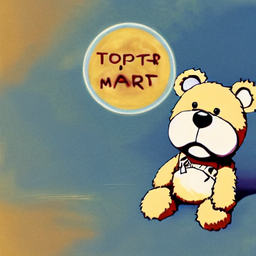} \\
             \includegraphics[height=2.3cm, width=2.3cm]{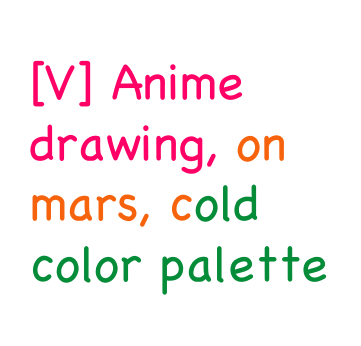} &  \includegraphics[height=2.3cm, width=2.3cm]{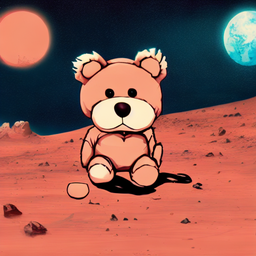} &  \includegraphics[height=2.3cm, width=2.3cm]{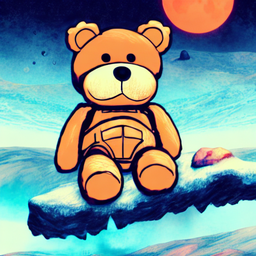} &   &  \includegraphics[height=2.3cm, width=2.3cm]{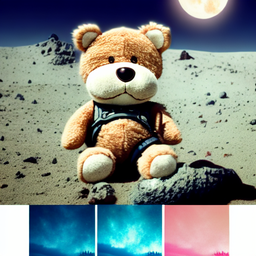} &  \includegraphics[height=2.3cm, width=2.3cm]{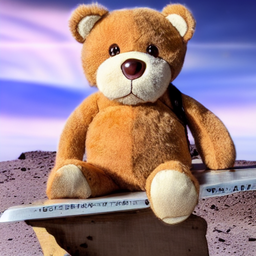} &  \includegraphics[height=2.3cm, width=2.3cm]{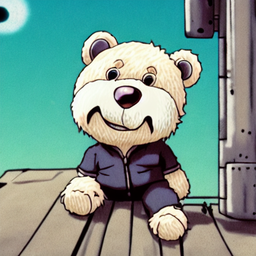} & \includegraphics[height=2.3cm, width=2.3cm]{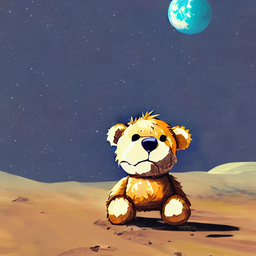} \\[4pt]

             \includegraphics[height=2.3cm, width=2.3cm]{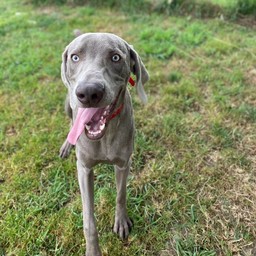} &  \includegraphics[height=2.3cm, width=2.3cm]{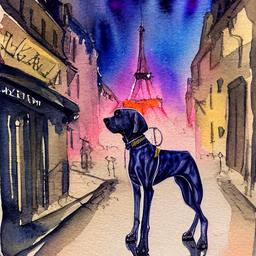} &  \includegraphics[height=2.3cm, width=2.3cm]{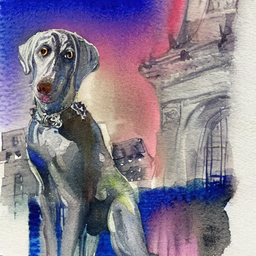} &   &  \includegraphics[height=2.3cm, width=2.3cm]{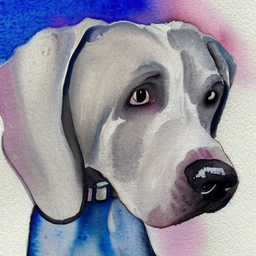} &  \includegraphics[height=2.3cm, width=2.3cm]{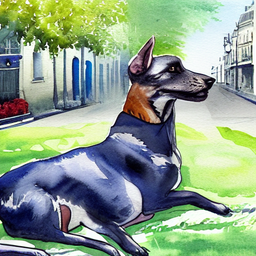} &  \includegraphics[height=2.3cm, width=2.3cm]{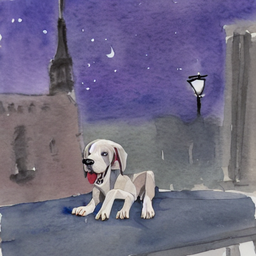} & \includegraphics[height=2.3cm, width=2.3cm]{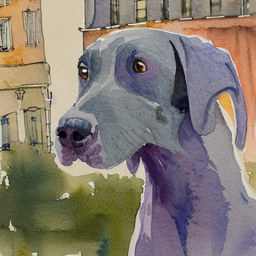} \\
             \includegraphics[height=2.3cm, width=2.3cm]{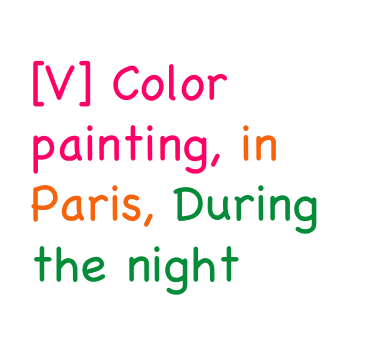} &  \includegraphics[height=2.3cm, width=2.3cm]{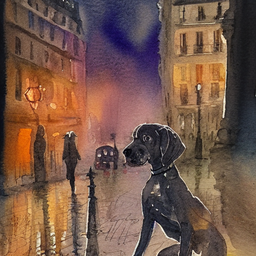} &  \includegraphics[height=2.3cm, width=2.3cm]{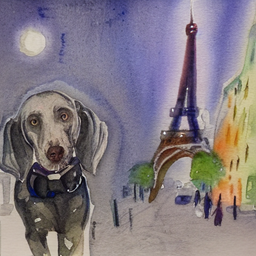} &   &  \includegraphics[height=2.3cm, width=2.3cm]{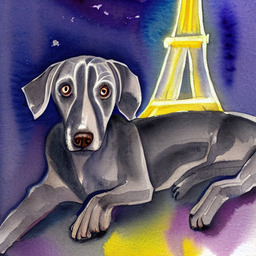} &  \includegraphics[height=2.3cm, width=2.3cm]{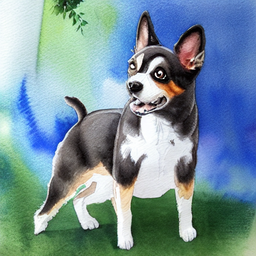} &  \includegraphics[height=2.3cm, width=2.3cm]{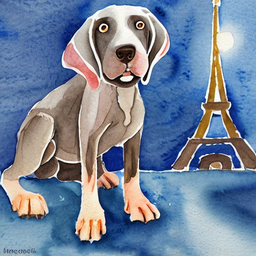} & \includegraphics[height=2.3cm, width=2.3cm]{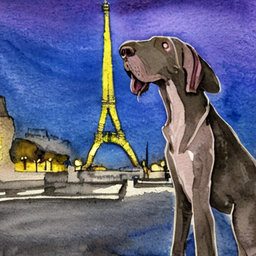} \\[4pt]

             \includegraphics[height=2.3cm, width=2.3cm]{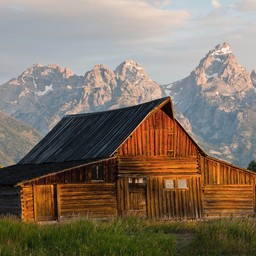} &  \includegraphics[height=2.3cm, width=2.3cm]{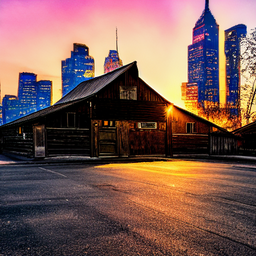} &  \includegraphics[height=2.3cm, width=2.3cm]{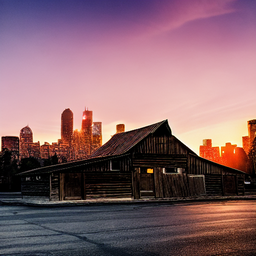} &   &  \includegraphics[height=2.3cm, width=2.3cm]{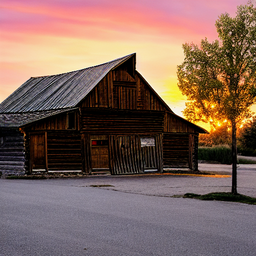} &  \includegraphics[height=2.3cm, width=2.3cm]{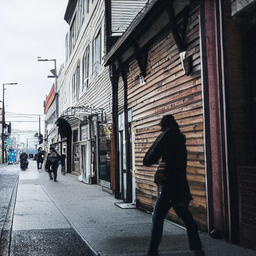} &  \includegraphics[height=2.3cm, width=2.3cm]{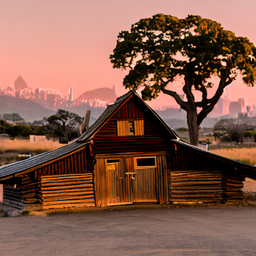} & \includegraphics[height=2.3cm, width=2.3cm]{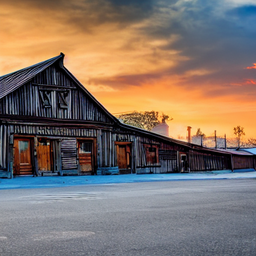} \\
             \includegraphics[height=2.3cm, width=2.3cm]{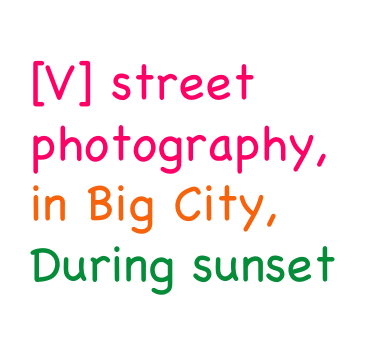} &  \includegraphics[height=2.3cm, width=2.3cm]{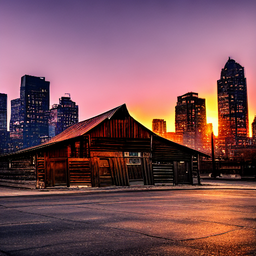} &  \includegraphics[height=2.3cm, width=2.3cm]{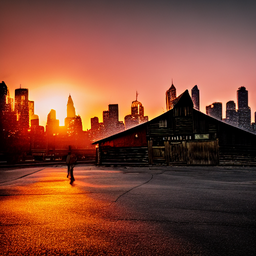} &   &  \includegraphics[height=2.3cm, width=2.3cm]{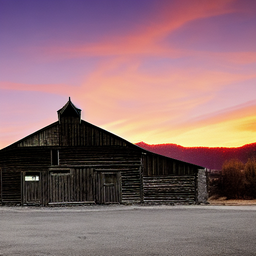} &  \includegraphics[height=2.3cm, width=2.3cm]{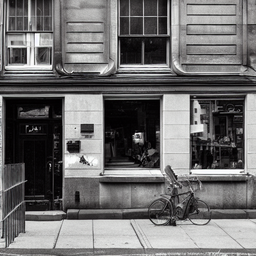} &  \includegraphics[height=2.3cm, width=2.3cm]{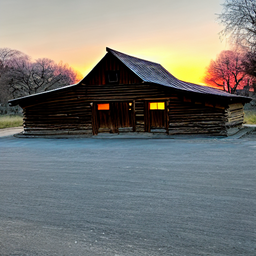} & \includegraphics[height=2.3cm, width=2.3cm]{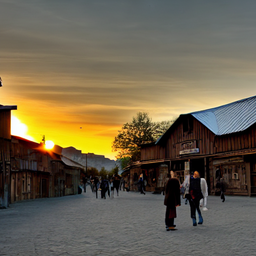} \\[4pt]

             \includegraphics[height=2.3cm, width=2.3cm]{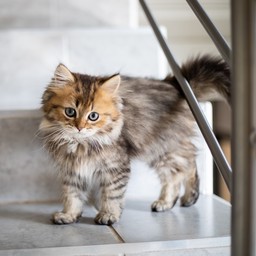} &  \includegraphics[height=2.3cm, width=2.3cm]{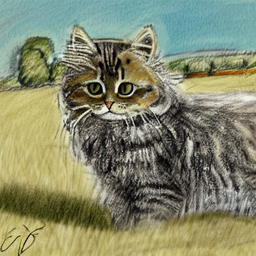} &  \includegraphics[height=2.3cm, width=2.3cm]{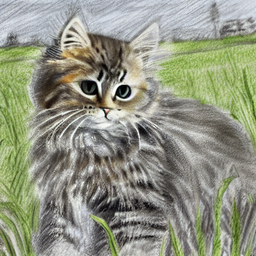} &   &  \includegraphics[height=2.3cm, width=2.3cm]{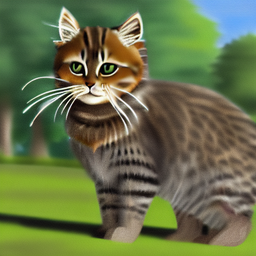} &  \includegraphics[height=2.3cm, width=2.3cm]{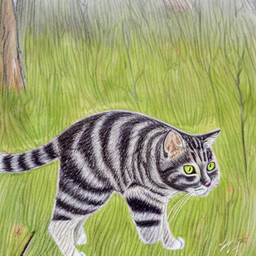} &  \includegraphics[height=2.3cm, width=2.3cm]{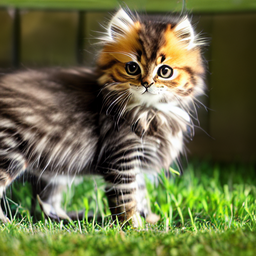} & \includegraphics[height=2.3cm, width=2.3cm]{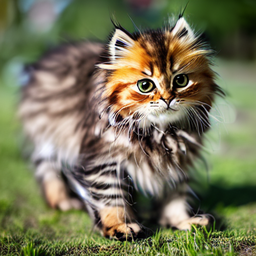} \\
             \includegraphics[height=2.3cm, width=2.3cm]{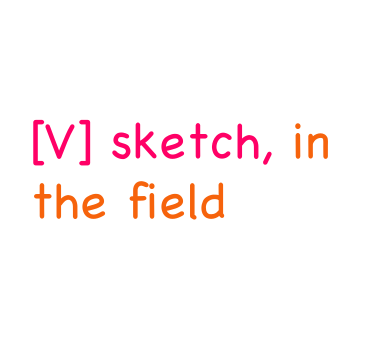} &  \includegraphics[height=2.3cm, width=2.3cm]{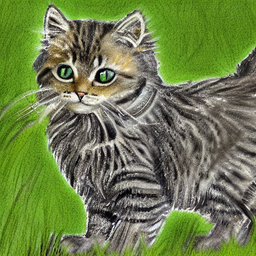} &  \includegraphics[height=2.3cm, width=2.3cm]{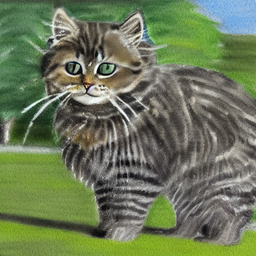} &   &  \includegraphics[height=2.3cm, width=2.3cm]{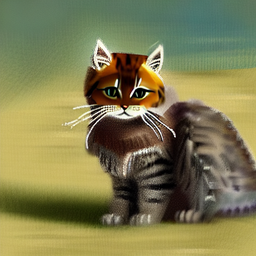} &  \includegraphics[height=2.3cm, width=2.3cm]{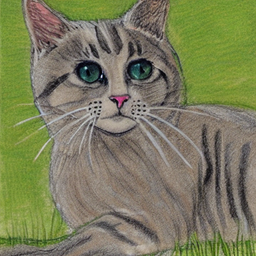} &  \includegraphics[height=2.3cm, width=2.3cm]{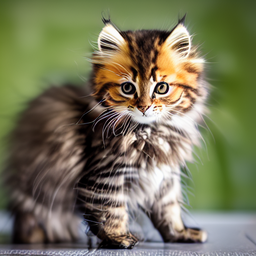} & \includegraphics[height=2.3cm, width=2.3cm]{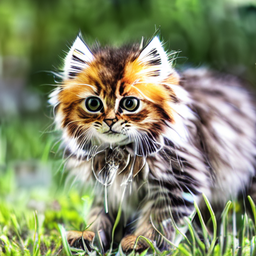} \\[4pt]
             
        \end{tabular}
\caption{Qualitative comparison in multi-shot setting. Our method achieves state-of-the-art results on complex prompts, better-preserving identity, and prompt alignment. For TI~\cite{TI}+DB~\cite{DB}, we use the same seed to generate our results, emphasizing the gain achieved by incorporating the prompt alignment path. For other baselines, we chose the best two out of eight samples.}
\label{fig:multishot_qualitative}
\end{table*}

\endgroup

\paragraph{Experimental setup:} We use StableDiffusion (SD)-v1.4~\cite{LDM} for ablation and comparison purposes, as many official implementations of state-of-the-art methods are available in SD-v1.4. We further validate our method with larger text-to-image models (see~\cref{sec:a_additional}). Complete experimental configuration, including learning rate, number of steps, and batch size, appear in ~\cref{sec:a_impl_details}. 

\paragraph{Evaluation metric: } for evaluation, we follow previous works~\cite{DB, TI} and use CLIP-score~\cite{CLIP} to measure alignment with the target (clean) prompt $y^c_t$ (i.e., does not have the placeholder [V]). For subject preservation, we also use CLIP feature similarity between the input and generated images with the target prompt. We use VIT-B32~\cite{VIT} trained by OpenAI on their proprietary data for both metrics. This ensures that the underlying CLIP used by SD-v1.4 differs from the one used for evaluation, which could compromise the validity of the reported metric.

\paragraph{Dataset: } for multi-shot setting, we use data collected by previous methods~\cite{TI, CustomDiffusion}, with different subjects like animals, toys, personal items, and buildings. For those subjects, checkpoints of previous methods exist, allowing a fair comparison. 

\subsection{Ablation studies}
\label{sec:ablation}
\begin{figure}
    \centering
    \includegraphics[width=0.95\linewidth]{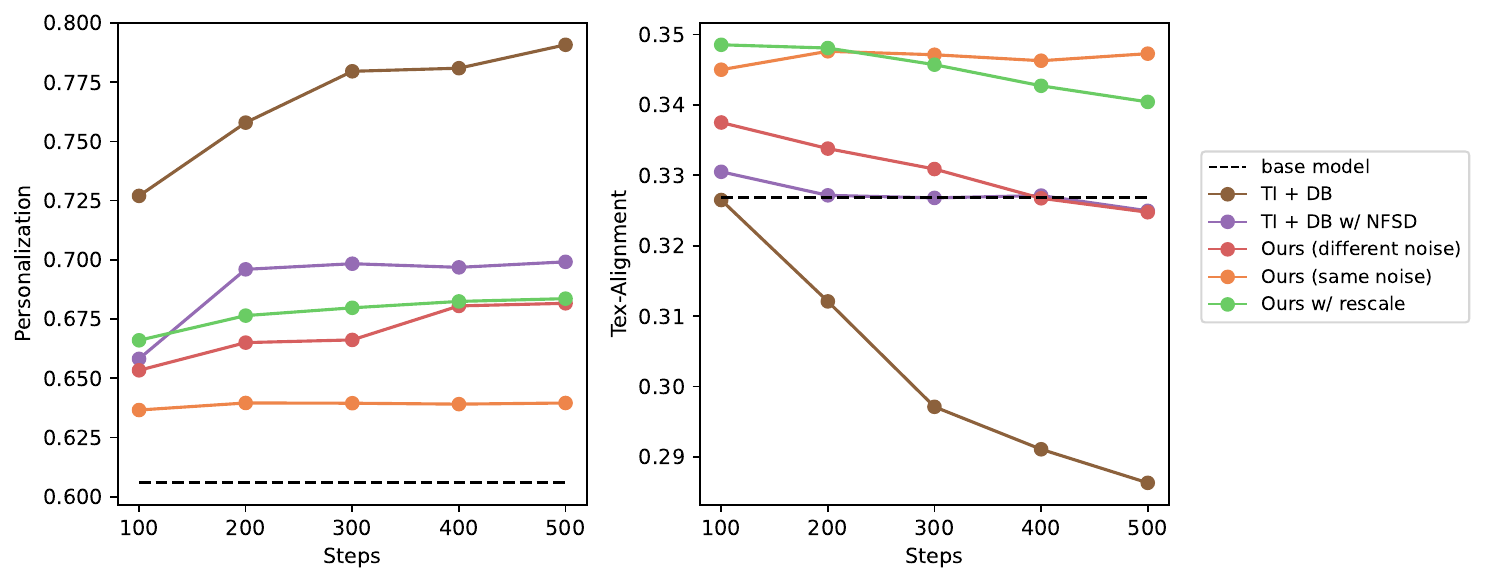}
    \caption{Ablation study. We report image alignment (left) and text alignment (right) as a function of the number of fine-tuning steps. The base model results are the pre-trained model's performance when using the class word to represent the target subject.}
    \label{fig:ablation}
\end{figure}
For ablation, we start with TI~\cite{TI} and DB~\cite{DB} as our baseline personalization method and gradually add different components contributing to our final method. 

\emph{\textbf{Early stopping: }} We begin by considering early stopping as a way to control the text-alignment. The lower the number of iterations, the less likely we are to hurt the model's prior knowledge. However, this comes at the cost of subject fidelity, evident from ~\cref{fig:ablation}. The longer we tune the model on the target subject, the more we risk overfitting the training set.

\emph{\textbf{Adding SDS guidance:}} improves text alignment, yet it severely harms the subject fidelity, and image diversity is substantially reduced (see ~\cref{sec:a_sds_vs_ours}). Alternative distillation sampling guidance~\cite{NFSD}  improves on top of SDS; however, since the distillation sampling guides the personalization optimization towards the center of distribution of the subject class, it still produces less favorable results.

 \emph{\textbf{Replacing SDS with PALP guidance:}}  improves text alignment by a considerable margin and maintains high fidelity to the subject $S$. We consider two variants: one where we use the same noise for personalization loss or sample a new one from the normal distribution. Interestingly, using the same noise helps with prompt alignment. Furthermore, scaling the score sampling equation~\cref{eq:palp_grad} by  $ \sqrt{\bar{\alpha}_{t}} / \sqrt{1-\bar{\alpha}_{t}}$ further enhances the performance.

\subsection{Comparison with Existing Methods}

\begin{table}
    \centering
    \resizebox{\columnwidth}{!}{
    
    \begin{tabular}{lcccccr}
    \toprule[1.5pt]
              & \multicolumn{5}{c}{Text-Alignment $ \uparrow $}&Image-Alignment $\uparrow$ \\
        Method& Style& Class& Ambiance 1& Ambiance 2& Target Prompt&\\[2pt]
        \midrule
         P+&  0.244   &  0.257&  0.217&  0.218&  0.308&0.673\\
         NeTI&  0.235 &  0.264&  0.22&  0.214&  0.310&0.695 \\
         TI+DB&  0.237 &  \textbf{0.279 }&  0.22&  0.216&  0.319&\textbf{0.716}\\
         Ours&  \textbf{0.245} &  0.272 &  \textbf{0.23} &  \textbf{0.224} &  \textbf{0.340} &0.681 \\
         \bottomrule[1.5pt]
    \end{tabular}
    }
    \caption{Comparisons to prior work. Our method presents better prompt-alignment, without hindering personalization. }
    \label{tab:multishot_quantitative_comp}
\end{table}
\begin{table}
    \centering
    \resizebox{0.7\columnwidth}{!}{

    \begin{tabular}{lcc}
        \toprule
              Method&  Text-Alignment $\uparrow$ & Personalization $ \uparrow $\\[2pt]
        \midrule
         P+&   68.5 \%           & 61.2 \% \\
         NeTI&   63.2 \%         &  70.3 \%\\
         TI+DB&   73.3 \%        &  60.4 \% \\
         Ours&   \textbf{91.2 \%} &\textbf{72.1 \%}\\
         \bottomrule
    \end{tabular}
    }
    \caption{User Study results. For text alignment, we report the percent of elements from the prompt that users found in the generated image. For personalization, users rated the similarity of the subject $S$ and the main subject in the generated image.}
    \label{tab:user_study}
\end{table}
\begingroup

\setlength{\tabcolsep}{2pt} %
\renewcommand{\arraystretch}{1.0} %

 \begin{table*}
        \centering
        \begin{tabular}{ccccccccc}
            Input & \hspace{5pt} & \multicolumn{3}{c}{Ours} & \hspace{5pt} & \multicolumn{3}{c}{Best of previous works}\\
             \includegraphics[width=0.125\textwidth, height=0.125\textwidth]{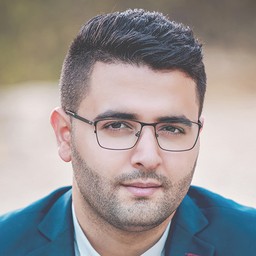} &  &  \includegraphics[width=0.125\textwidth, height=0.125\textwidth]{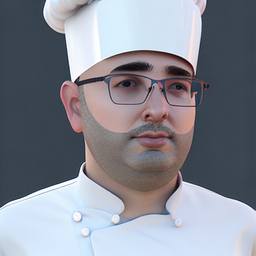}  & \includegraphics[width=0.125\textwidth, height=0.125\textwidth]{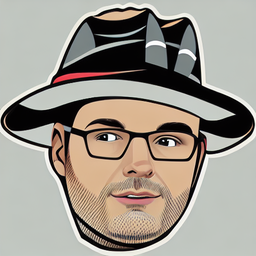} &  \includegraphics[width=0.125\textwidth, height=0.125\textwidth]{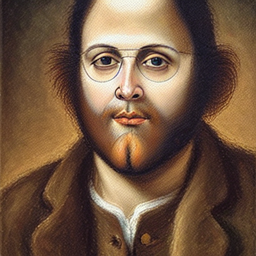} & & \includegraphics[width=0.125\textwidth, height=0.125\textwidth]{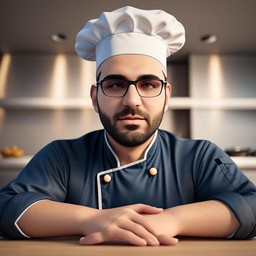} & \includegraphics[width=0.125\textwidth, height=0.125\textwidth]{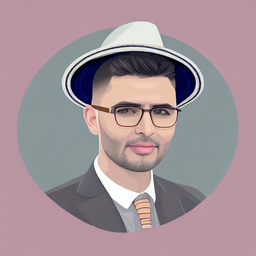} &  \includegraphics[width=0.125\textwidth, height=0.125\textwidth]{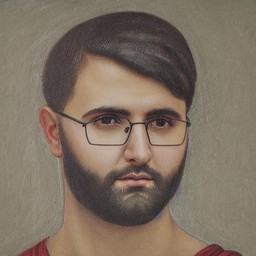}\\
              & & \makecell{"3D Render \\ as a chef"}  & \makecell{"Vector art \\ wearing a hat"} & \makecell{"A painting \\ by Da Vinci"} & & \makecell{"3D Render \\ as a chef"}  & \makecell{"Vector art \\ wearing a hat"} & \makecell{"A painting \\ by Da Vinci"}\\

              & &  \includegraphics[width=0.125\textwidth, height=0.125\textwidth]{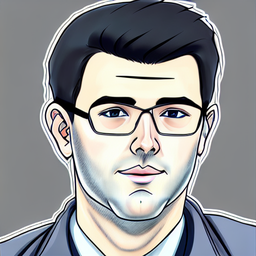}  & \includegraphics[width=0.125\textwidth, height=0.125\textwidth]{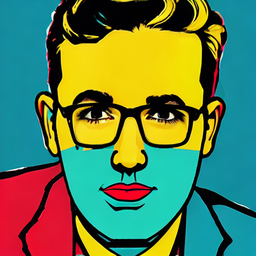} &  \includegraphics[width=0.125\textwidth, height=0.125\textwidth]{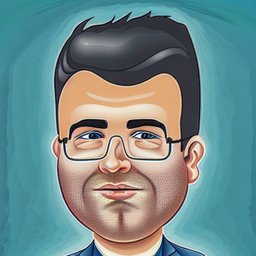} & & \includegraphics[width=0.125\textwidth, height=0.125\textwidth]{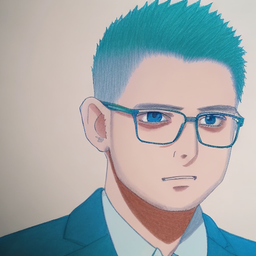} & \includegraphics[width=0.125\textwidth, height=0.125\textwidth]{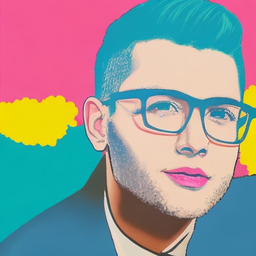} &  \includegraphics[width=0.125\textwidth, height=0.125\textwidth]{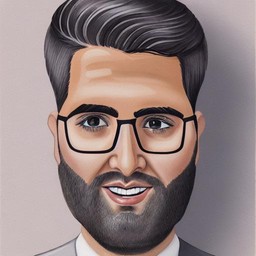}\\
              & & \makecell{"Anime drawing"}  & \makecell{"Pop art"} & \makecell{"A Caricature"} & & \makecell{"Anime drawing"}  & \makecell{"Pop art"} & \makecell{"A Caricature"} \\
              
              \vspace{4pt} \\

             \includegraphics[width=0.125\textwidth, height=0.125\textwidth]{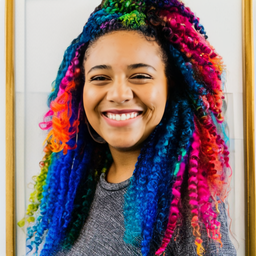} &  &  \includegraphics[width=0.125\textwidth, height=0.125\textwidth]{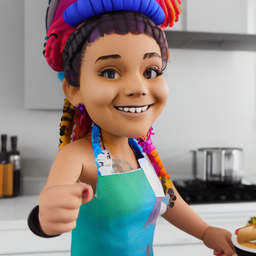}  & \includegraphics[width=0.125\textwidth, height=0.125\textwidth]{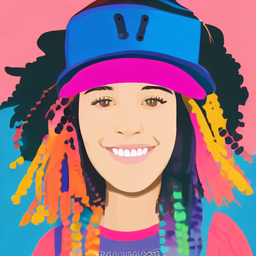} &  \includegraphics[width=0.125\textwidth, height=0.125\textwidth]{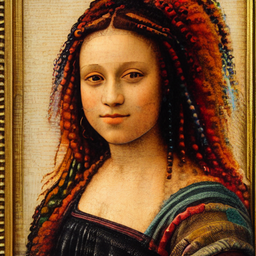} & & \includegraphics[width=0.125\textwidth, height=0.125\textwidth]{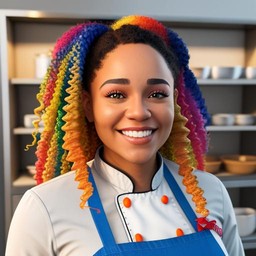} & \includegraphics[width=0.125\textwidth, height=0.125\textwidth]{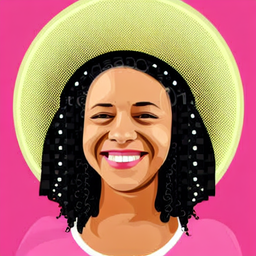} &  \includegraphics[width=0.125\textwidth, height=0.125\textwidth]{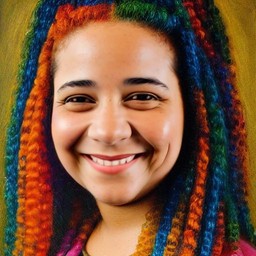}\\
              & & \makecell{"3D Render \\ as a chef"}  & \makecell{"Vector art \\ wearing a hat"} & \makecell{"A painting \\ by Da Vinci"} & & \makecell{"3D Render \\ as a chef"}  & \makecell{"Vector art \\ wearing a hat"} & \makecell{"A painting \\ by Da Vinci"}\\

              & &  \includegraphics[width=0.125\textwidth, height=0.125\textwidth]{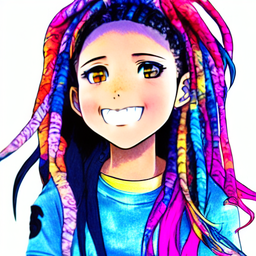}  & \includegraphics[width=0.125\textwidth, height=0.125\textwidth]{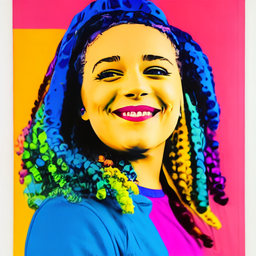} &  \includegraphics[width=0.125\textwidth, height=0.125\textwidth]{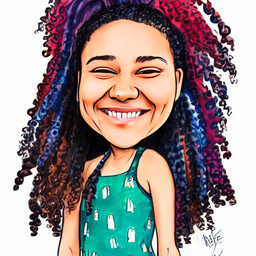} & & \includegraphics[width=0.125\textwidth, height=0.125\textwidth]{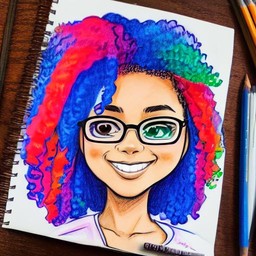} & \includegraphics[width=0.125\textwidth, height=0.125\textwidth]{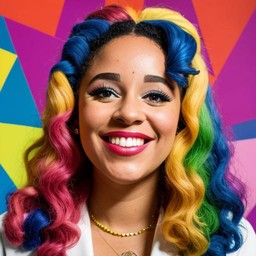} &  \includegraphics[width=0.125\textwidth, height=0.125\textwidth]{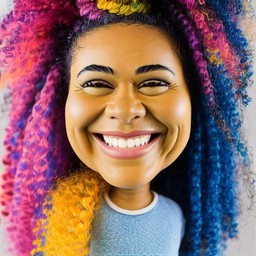}\\
              & & \makecell{"Anime drawing"}  & \makecell{"Pop art"} & \makecell{"A Caricature"} & & \makecell{"Anime drawing"}  & \makecell{"Pop art"} & \makecell{"A Caricature"} \\
              
        \end{tabular}
        \caption{Qualitative comparison against ProFusion~\cite{ProFusion}, IP-Adapter~\cite{IPAdapter} , E4T~\cite{E4T}, and Face0~\cite{Face0}. On the left, we show the results of our method on two individuals using a \emph{single} image with multiple prompts. To meet space requirements, we report a single result among all previous methods, based on our preference, for a  given subject and prompt. Full comparison appears in ~\cref{fig:sup_faces_full_guy}~\cref{fig:sup_faces_full_girl}.}

        \label{fig:faces_qualitative}
\end{table*}

\endgroup

We compare our method against multi-shot methods, including CustomDiffusion~\cite{CustomDiffusion}, $P+$~\cite{PPLUS}, and NeTI~\cite{NeTI}. We further compare against TI~\cite{TI} and DB~\cite{DB} using our implementation, which should also highlight the gain we achieve by incorporating our framework with existing personalization methods. Our evaluation set contains ten different complex prompts that include at least four different elements, including style-change (e.g., "sketch of", "anime drawing of"), time or a place (e.g., "in Paris", "at night"), color palettes (e.g., "warm," "vintage"). We asked users to count the number of elements that appear in the image (text-alignment) and rate the similarity of the results to the input subject (personalization, see ~\cref{tab:multishot_quantitative_comp}).

Our method achieves the best text alignment while maintaining high image alignment. TI+DB achieves the best image alignment. However, the reason for this is because TI+DB is prone to over-fitting. Indeed, investigating each element in the prompt, we find that the TI+DB achieves the best alignment with the class prompt (e.g., "A  photo of a cat") while being significantly worse in the Style prompt (e.g., "A sketch"). Our method has a slightly worse image alignment since we expect appearance change for stylized prompts. We validate this hypothesis with a user study and find that our method achieves the best user preference in prompt alignment and personalization (see~\cref{tab:user_study}). Full details on the user study appear in~\cref{sec:a_user_study}.

\subsection{Applications}

\paragraph{Single-shot setting:} In a single-shot setting, we aim to personalize text-2-image models using a single image. This setting is helpful for cases where only a single image exists for our target subject (e.g., an old photo of a loved one). For this setting, we qualitatively compare our method with encoder-based methods, including IP-Adapter~\cite{IPAdapter}, ProFusion~\cite{ProFusion}, Face0~\cite{Face0}, and E4T~\cite{E4T}. We use portraits of two individuals and expect previous methods to generalize to our selected images since all methods are pre-trained on human faces. Note that E4T~\cite{E4T} and ProFusion~\cite{ProFusion} also perform test time optimization.

As seen from~\cref{fig:faces_qualitative}, our method is both prompt- and identity-aligned. Previous methods, on the other hand, struggle more with identity preservation. We note that optimization-based approaches~\cite{E4T, ProFusion} are more identity-preserving, but this comes at the cost of text alignment. Finally, our method achieves a higher success rate, where the quality of the result is independent of the chosen seed.

\paragraph{Multi-concept Personalization: } Our method accommodates multi-subject personalization via simple modifications. Assume we want to compose two subjects, $S_1$ and $S_2$, in a specific scene depicted by a given prompt $y$. To do so, we first allocate two different placeholders, [V1] and [V2], to represent the target subjects $S_1$ and $S_2$, respectively. During training, we randomly sample an image from a set of images containing $S_1$ and $S_2$. We assign different personalization prompts $y_P$ for each subject, e.g., "A photo of [V1]" or "A painting inspired by [V2]", depending on the context. Then, we perform PALP while using the target prompt in mind, e.g., "A painting of [V1] inspired by [V2]". This allows composing different subjects into coherent scenes or using a single artwork as a reference for generating art-inspired images. Results appear in~\cref{fig:composition}; further details and results appear in~\cref{sec:a_multisubject}. 
\section{Conclusions}
We have introduced a novel personalization method that allows better prompt alignment. Our approach involves fine-tuning a pre-trained model to learn a given subject while employing score sampling to maintain alignment with the target prompt. We achieve favorable results in both prompt- and subject-alignment and push the boundary of personalization methods to handle complex prompts, comprising multiple subjects, even when one subject has only a single reference image.

While the resulting personalized model still generalizes for other prompts, we must personalize the pre-trained model for different prompts to achieve optimal results. For practical real-time use cases, there may be better options. However, future directions employing prompt-aligned adapters could result in instant time personalization for a specific prompt (e.g., for sketches). Finally, our work will motivate future methods to excel on a subset of prompts, allowing more specialized methods to achieve better and more accurate results.

{
    \small
    \bibliographystyle{ieeenat_fullname}
    \bibliography{main}
}

\appendix
\section{Over-saturation and mode-collapse in SDS}
\label{sec:a_sds_vs_ours}
\begin{figure}
    \centering
    \includegraphics[width=\linewidth]{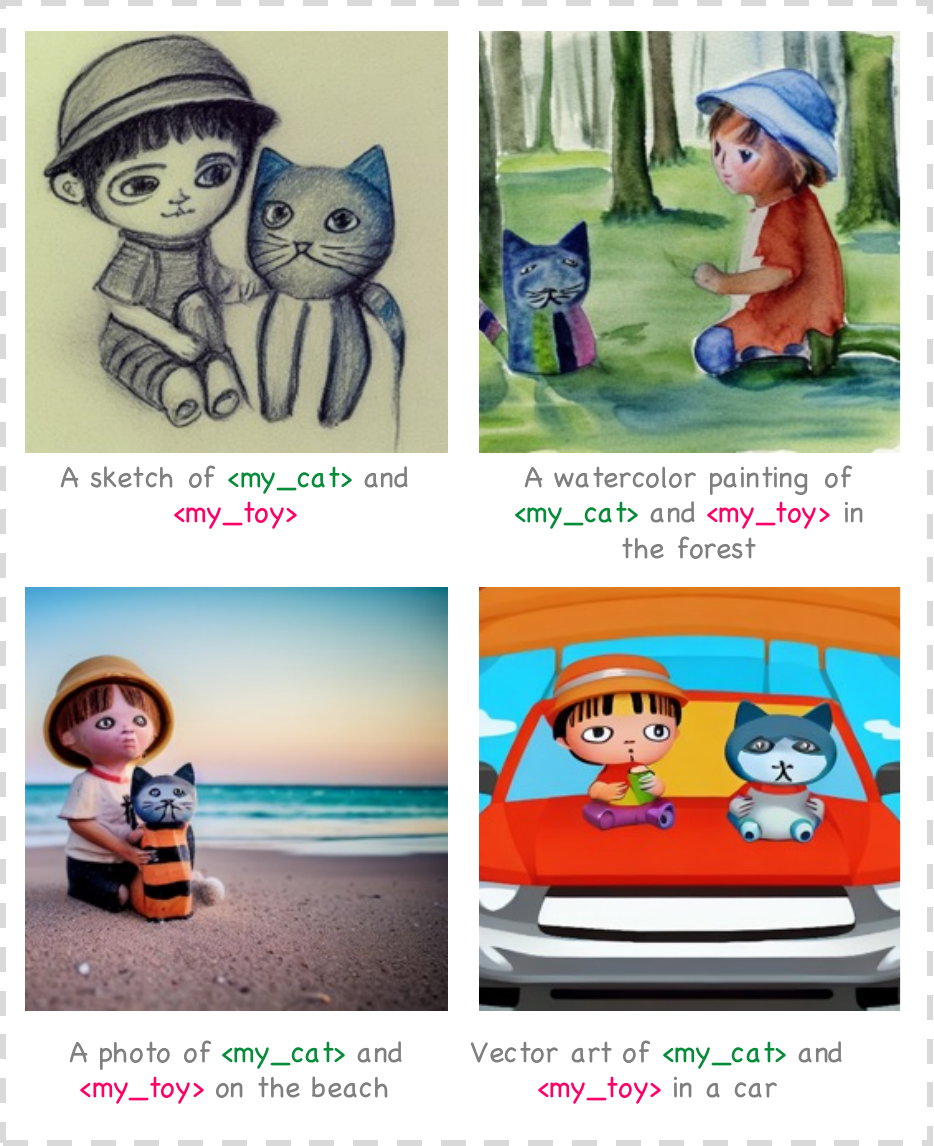}
    \caption{Additional results for Multi-subject personalization. The reference images appear in ~\cref{fig:art_fig}.}
    \label{fig:additional_composition}
\end{figure}
\begin{figure}
    \centering
    \frame{\includegraphics[width=0.45\linewidth]{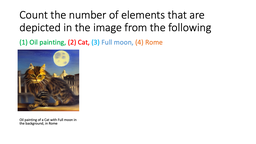}}
    \hspace{4pt}
    \frame{\includegraphics[width=0.45\linewidth]{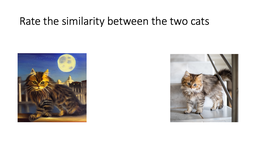}}
    \caption{User study sample questions. On the left, we show a sample question in which the participants must count the prompt elements perceived from the generated sample. On the right, users were asked to rate, from 1 to 5, the similarity between the two main subjects, with one being the least similar. Participants answers were aggregated and normalized.}
    \label{fig:user_study_sample_questions}
\end{figure}
\begin{figure*}
    \centering
    \begin{subfigure}{0.33\textwidth}
        \centering
        \includegraphics[width=0.45\textwidth]{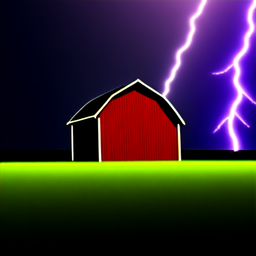}
        \includegraphics[width=0.45\textwidth]{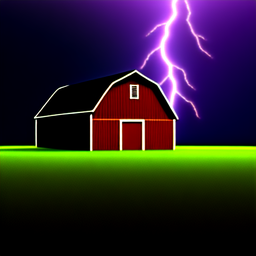}
        \includegraphics[width=0.45\textwidth]{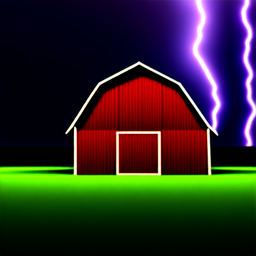}
        \includegraphics[width=0.45\textwidth]{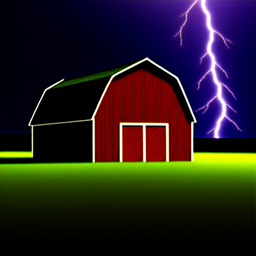}
        \caption{SDS~\cite{SDS}}    
    \end{subfigure}
    \begin{subfigure}{0.33\linewidth}
        \centering
        \includegraphics[width=0.45\textwidth]{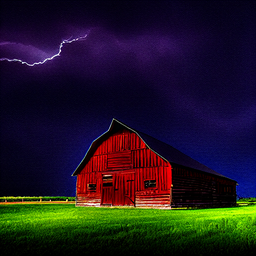}
        \includegraphics[width=0.45\textwidth]{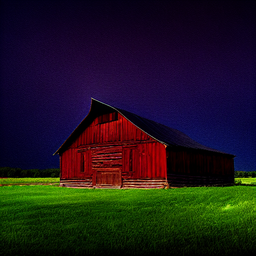}
        \includegraphics[width=0.45\textwidth]{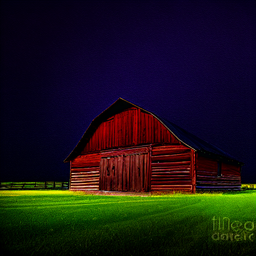}
        \includegraphics[width=0.45\textwidth]{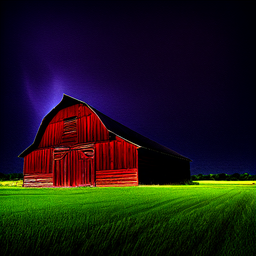}
        \caption{NFSD~\cite{NFSD}}    
    \end{subfigure}
    \begin{subfigure}{0.33\linewidth}
        \centering
        \includegraphics[width=0.45\textwidth]{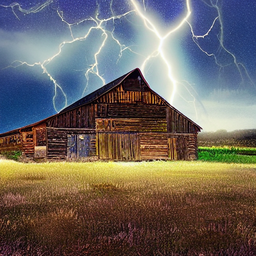}
        \includegraphics[width=0.45\textwidth]{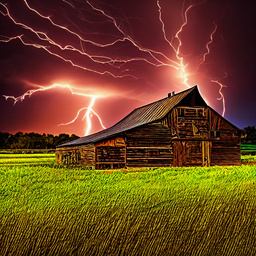}
        \includegraphics[width=0.45\textwidth]{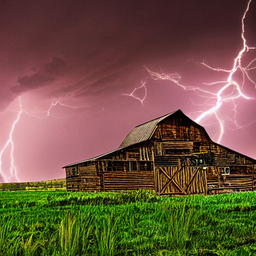}
        \includegraphics[width=0.45\textwidth]{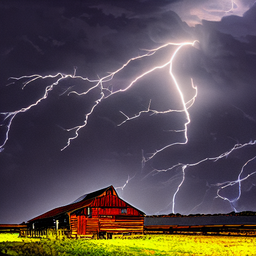}
        \caption{Ours}    
    \end{subfigure}
    
    \label{fig:sds_vs_ours}
    \caption{Using SDS~\cite{SDS} guidance for prompt alignment produces over-saturated results. Alternative improvements like NFSD~\cite{NFSD} producess less-diverse results. Our method produces diverse and prompt-aligned results. The input prompt is "A digital art of [V] on a field with lightning in the background."}
\end{figure*}

We use prompt-aligned score sampling~\cref{eq:palp_grad} to guide the model towards the target prompt $y$. We found this to be more effective than using SDS~\cite{SDS} or improved versions of it like NFSD~\cite{NFSD}. In particular, SDS and NFSD produce over-saturated or less diverse results. We provide qualitative examples in~\cref{fig:sds_vs_ours}, and quantitative comparison appears in~\cref{sec:ablation}.

\section{Additional Results}
\label{sec:a_additional}
\begin{figure*}
    \centering
    \includegraphics[width=0.75\textwidth]{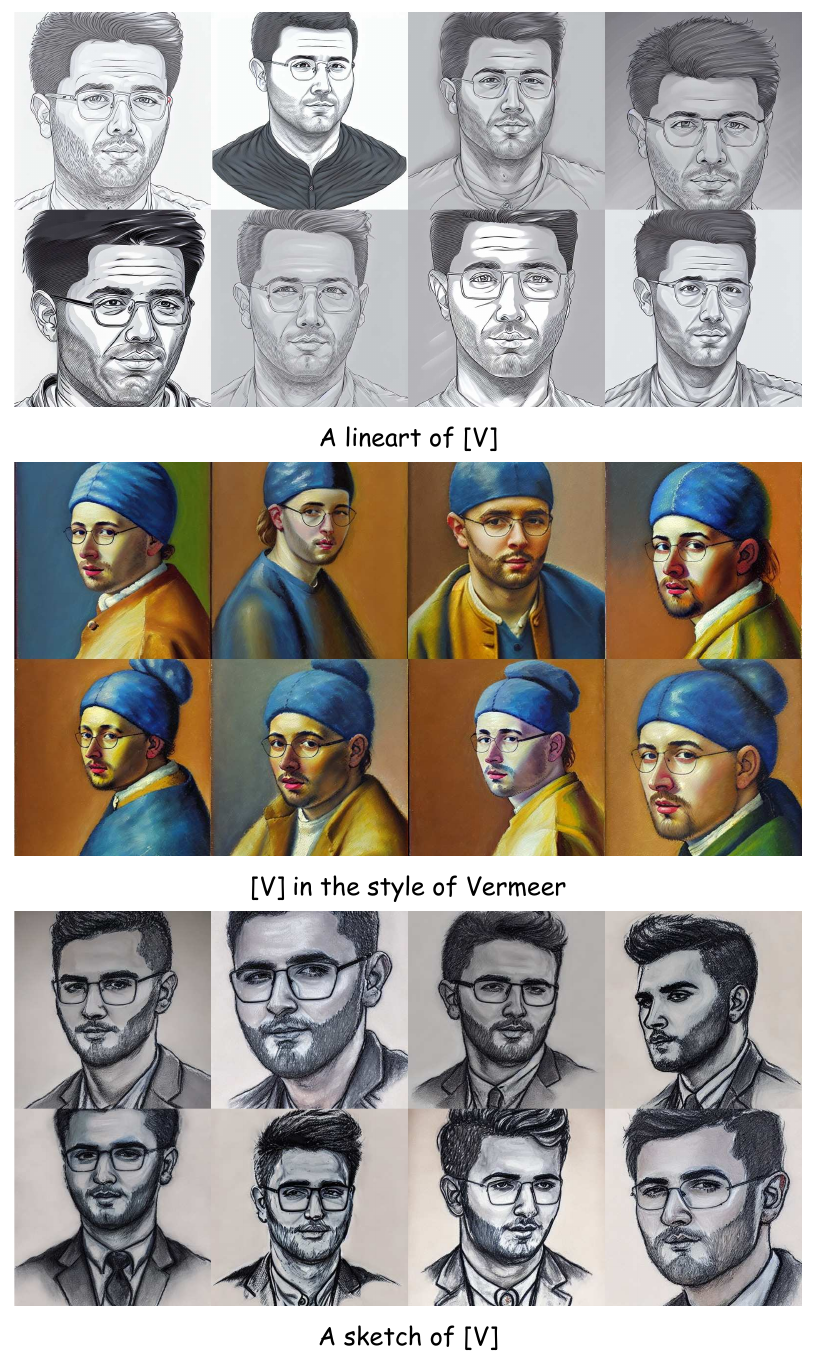}
    \caption{Un-curated samples for single-shot setting. The input image appear in~\cref{fig:sup_faces_full_guy}.}
    \label{fig:uncurated_single_shot}
\end{figure*}

We provide additional qualitative comparison for multi-shot setting, the results appear in~\cref{fig:multishot_qualitative_additional}. Moreover, a full qualitative comparison and un-curated results of the single-shot experiment appear in~\cref{fig:sup_faces_full_girl}, ~\cref{fig:sup_faces_full_guy}, and~\cref{fig:uncurated_single_shot}. For results obtained using a larger model, with improved text-encoder and bigger capacity, see~\cref{tab::LARGE_1} and~\cref{tab:LARGE_2}.

\section{Implementation Details}
\label{sec:a_impl_details}
All our experiments, except those conducted using a larger model, were conducted using TPU-v3, with a global batch size of 32. We use a learning rate of $\texttt{5e-5}$. For LoRA~\cite{LORA}, we use a rank $r=32$ and only modify the self-and cross-attention layers' projection matrices. For most experiments, we use a classifier free-guidance scale of $\alpha=15.0$ and $\alpha=7.5$; for composition experiments, we use $\alpha=7.5$ and $\beta=1.0$. For quantitative comparison, we fine-tuned the model for 500 steps. However, this may be sub-optimal depending on the subject and prompt complexity.

\section{User-Study Details}
\label{sec:a_user_study}
In the user study, we asked 30 individuals to rate prompt alignment and personalization performances of four methods, including $P+$~\cite{PPLUS}, NeTI~\cite{NeTI}, TI~\cite{TI}+DB~\cite{DB}, and our method. Our test set includes six different subjects and ten prompts. We generated eight samples for each prompt and subject using the four methods. Then, we randomly picked a single photo and asked the participants to count the number of different prompt elements that appear in the generated image. We asked the users to rate the similarity between the main subject in the generated image and the sample. Then, we randomly divided the questions into three forms and shuffled them between the participants. Sample questions appear in~\cref{fig:user_study_sample_questions}.

\section{Multi-subject Personalization}
\label{sec:a_multisubject}
\begin{figure*}
    \centering
    \includegraphics[width=\textwidth]{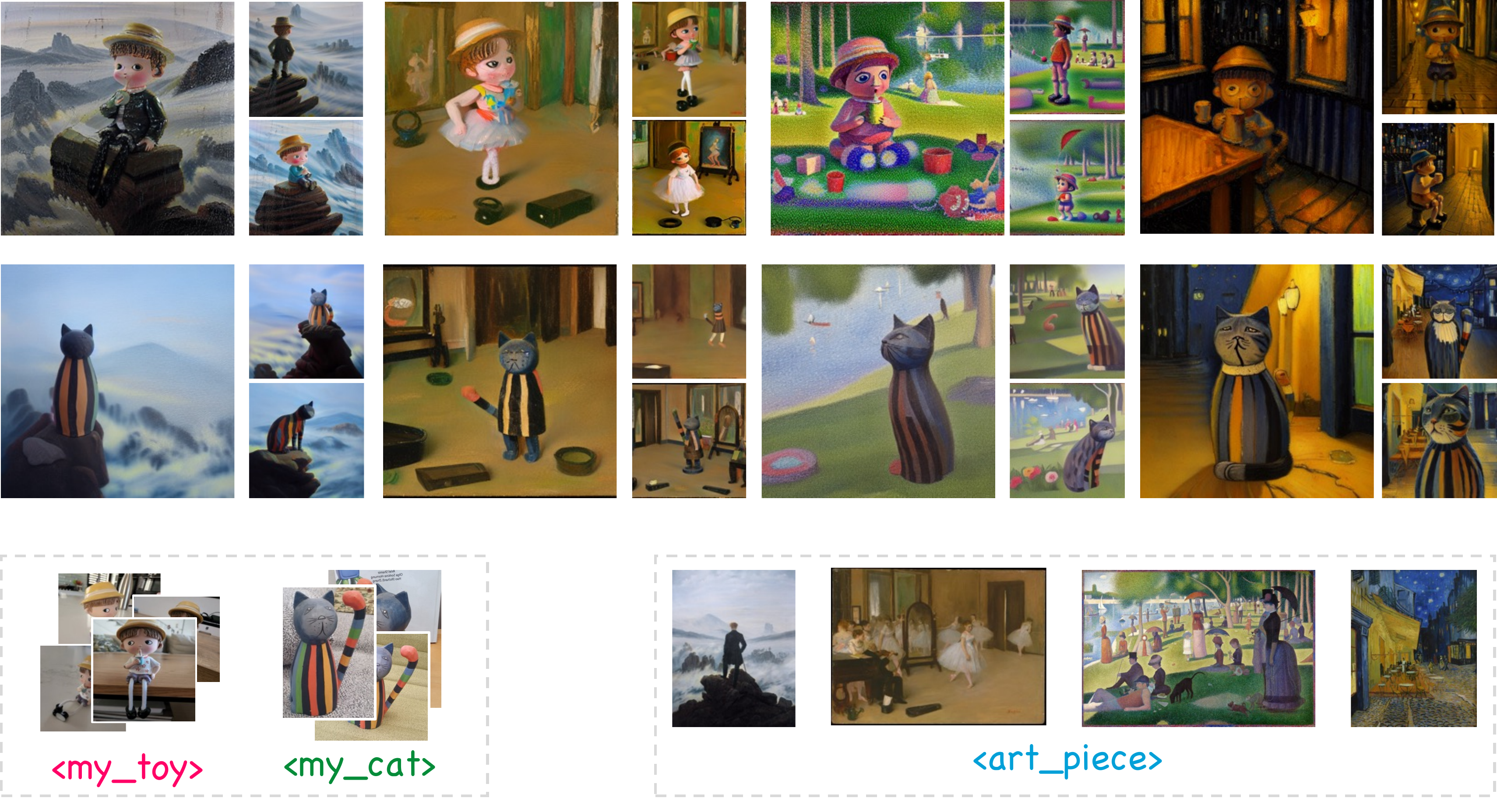}
    \caption{Art inspired composition. Our method blends the target subject and the reference paintings coherently. Further, we can produce diverse results by slightly modifying the prompt. For example, we can generate an image of $\texttt{<my toy>}$ having a picnic, playing with a kite, or simply standing next to a lake (see top-row third column).}
    \label{fig:art_fig}
\end{figure*}
\begin{figure*}
    \centering
    \begin{subfigure}{0.33\textwidth}
        \centering
        \includegraphics[width=0.9\textwidth]{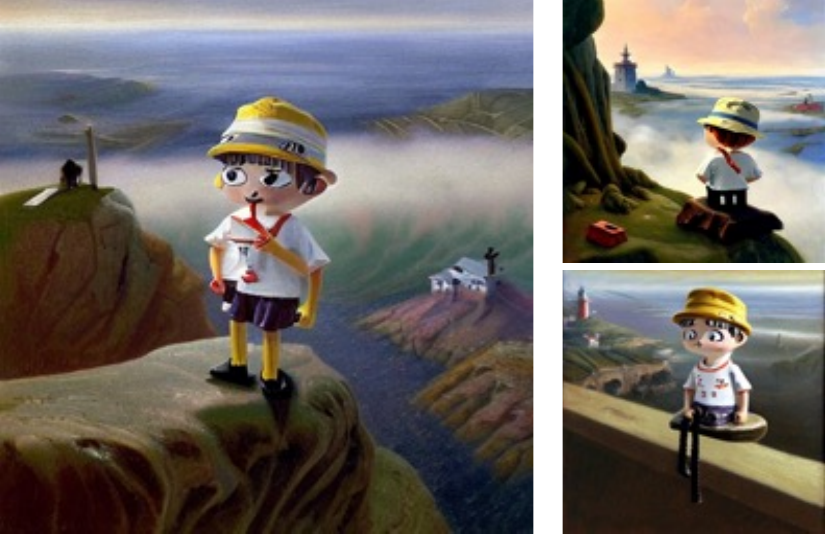}
        \caption{NeTI~\cite{NeTI}}    
    \end{subfigure}
    \begin{subfigure}{0.33\linewidth}
        \centering
        \includegraphics[width=0.9\textwidth]{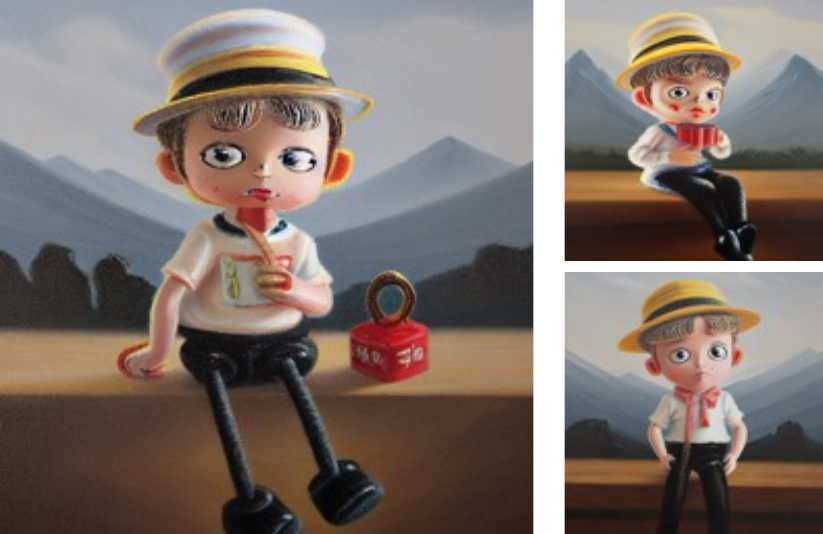}
        \caption{w/o PALP Guidnace}    
    \end{subfigure}
    \begin{subfigure}{0.33\linewidth}
        \centering
        \includegraphics[width=0.9\textwidth]{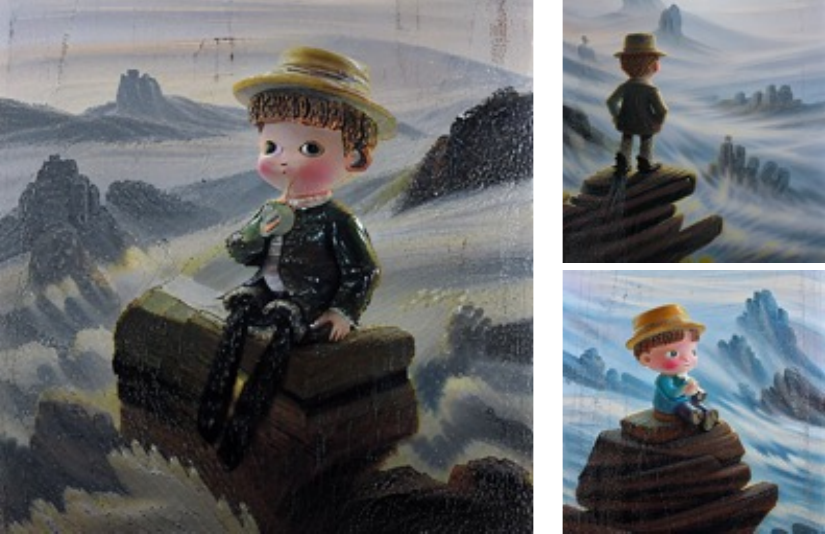}
        \caption{Ours}    
    \end{subfigure}
    \caption{Art-inspired composition ablation. We consider two alternatives: (a) using a pre-trained personalized model and prompting the artwork name, or (b) jointly training on both subjects without PALP guidance. In both cases, the results are sub-optimal, where our method achieves more coherent results. Reference images appear in~\cref{fig:art_fig}.}
    \label{fig:art_ablation}
\end{figure*}

\paragraph{Multi-subject composition:} For multi-subject personalization, we use two placeholder $V_1$ and $V_2$ to represent the subject $S_1$ and $S_2$. We use $y_P$ = "A photo of [$V_i$]" as our personalization prompt. We use a descriptive prompt describing the desired scene for the target clean prompt. For example, "A 19th century portrait of a [$C_1$] and [$C_2$]", where $C_1$ and $C_2$ represent the class name of $S_1$ and $S_2$, respectively. We provide additional results in~\cref{fig:additional_composition}.

\paragraph{Art-inspired composition:} For Art-inspired composition, we use "A photo of [$V_1$]" to describe the main subject, while "An oil painting of [$V_2$]" is used for the artistic image. The clean prompt used is "An oil painting of [class name]," where "[class name]" is the subject class (e.g., a cat or a toy). Furthermore, adding a description to the clean prompt improves alignment, for example, "An oil painting of a cat sitting on a rock" for the" Wanderer above the Sea of Fog" artwork by Caspar David Friedrich. Finally, at test time, we use the target prompt "An oil painting of [$V_1$] inspired by [$V_2$] painting" or a similar variant, with possibly an additional description of the desired scene. Results appear in~\cref{fig:art_fig}. We further show that joint training is insufficient for this task, nor is pre-training on $S$ prompting the artwork or artist name. Both cases produce miss-aligned results (see~\cref{fig:art_ablation}).

As can be seen from~\cref{fig:methodvisx0hat}, the pre-trained model (before personalization) steers the image towards a sketch-like image, where the estimate $\hat{x}_0$ has a white background, but the subject details are missing. Apply personalization without PALP overfits, where elements from the training images, like the background, are more dominant, and sketchiness fades away, suggesting miss-alignment with the prompt ``A sketch'' and over-fitting to the input image. Using PALP, the model prediction steers the backward-denoising process towards a sketch-like image, while staying personalized, where a cat like shape is restored.

\begin{table*}
    \centering
    \begin{tabular}{cc}
         w/o PALP&  w/ PALP\\[4pt]
         \includegraphics[width=0.45\textwidth]{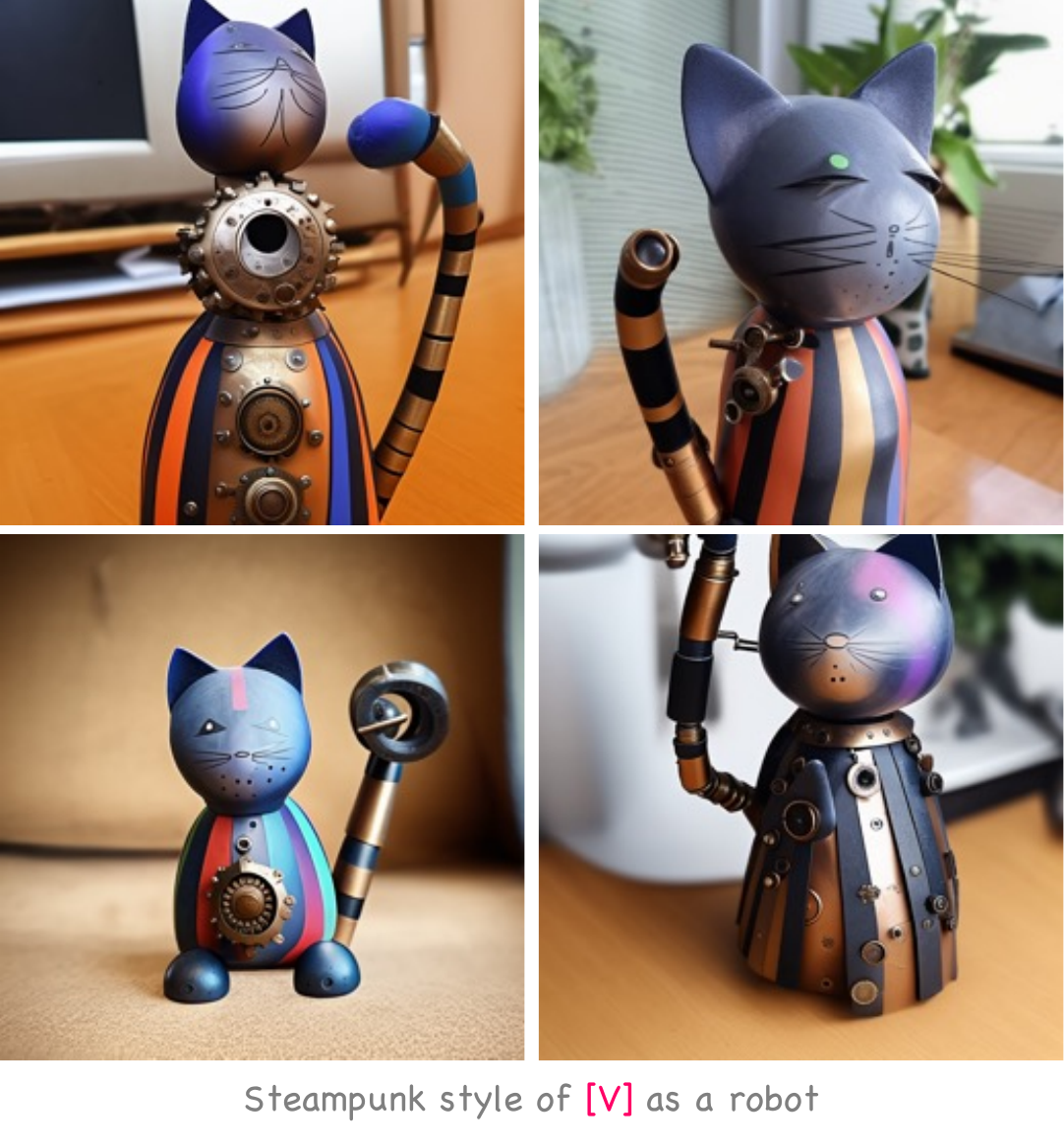}& \includegraphics[width=0.45\textwidth]{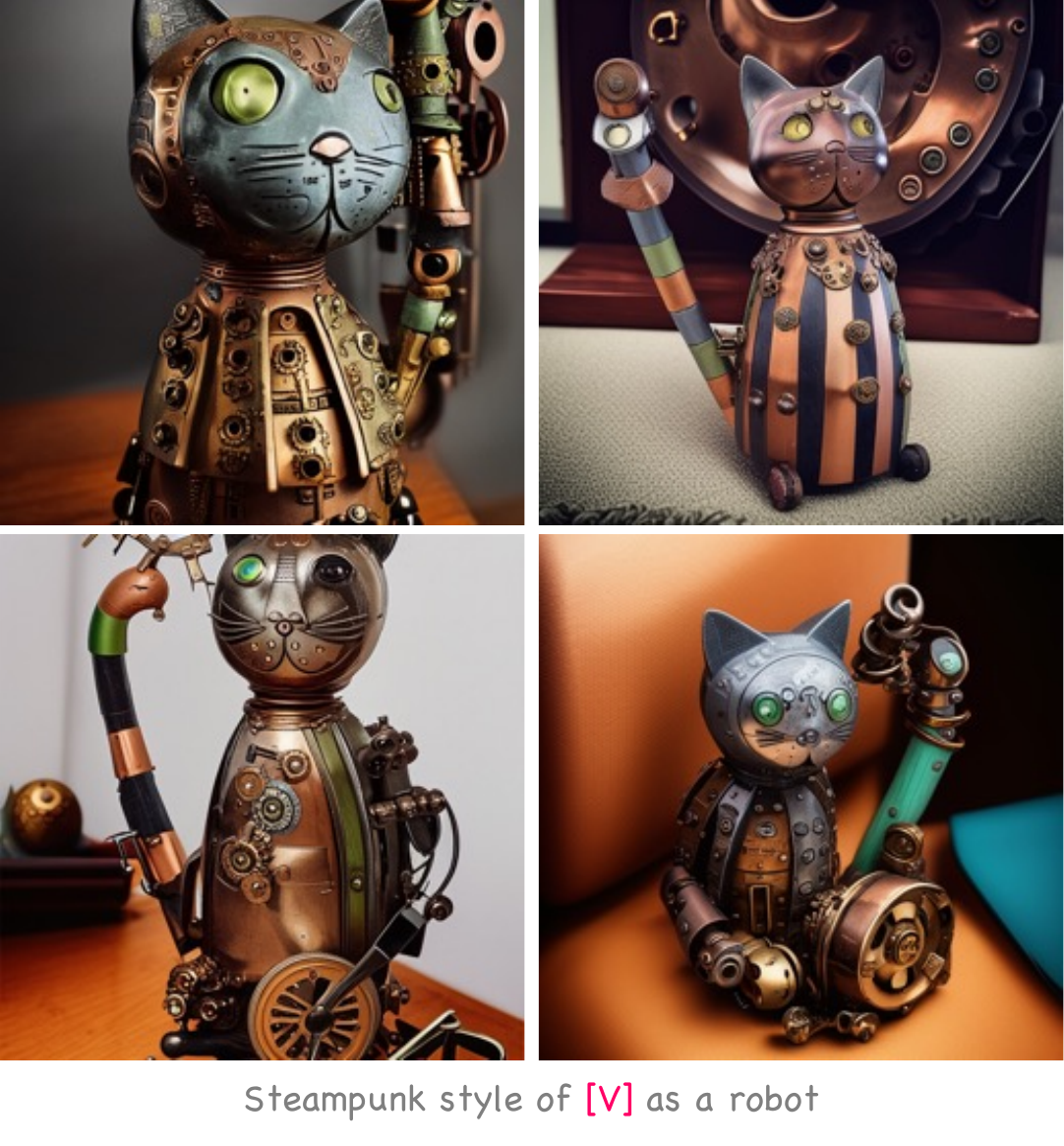} \\
         \includegraphics[width=0.45\textwidth]{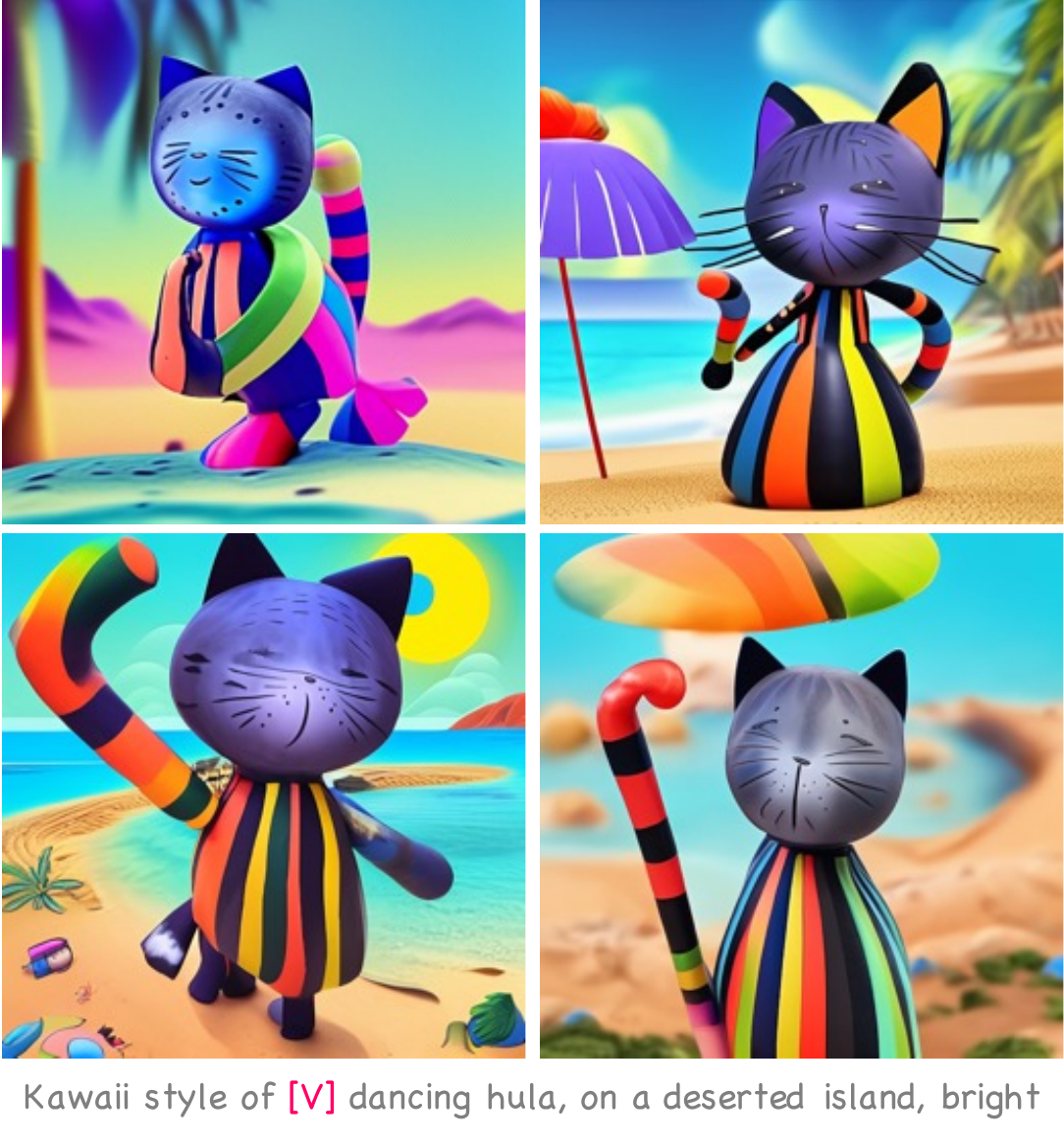} & \includegraphics[width=0.45\textwidth]{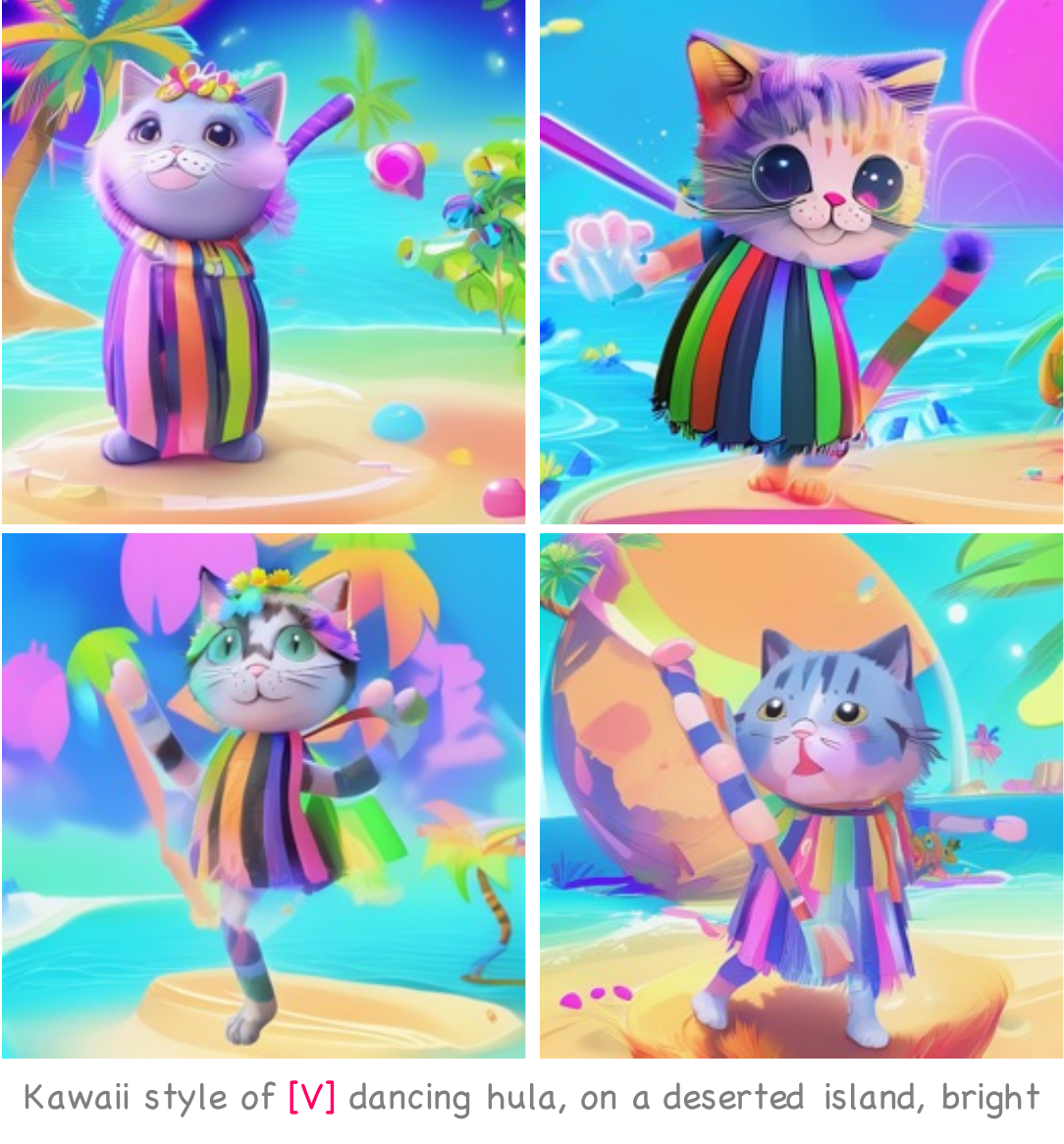}\\
         \multicolumn{2}{c}{\includegraphics[width=0.95\textwidth]{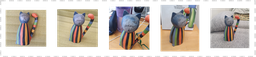}}
    \end{tabular}
    \caption{Additional Results on Larger Text-to-Image Models. Reference images appear at the bottom.}
    \label{tab::LARGE_1}
\end{table*}
\begin{table*}
    \centering
    \begin{tabular}{cc}
         w/o PALP&  w/ PALP\\[4pt]
        \includegraphics[width=0.45\textwidth]{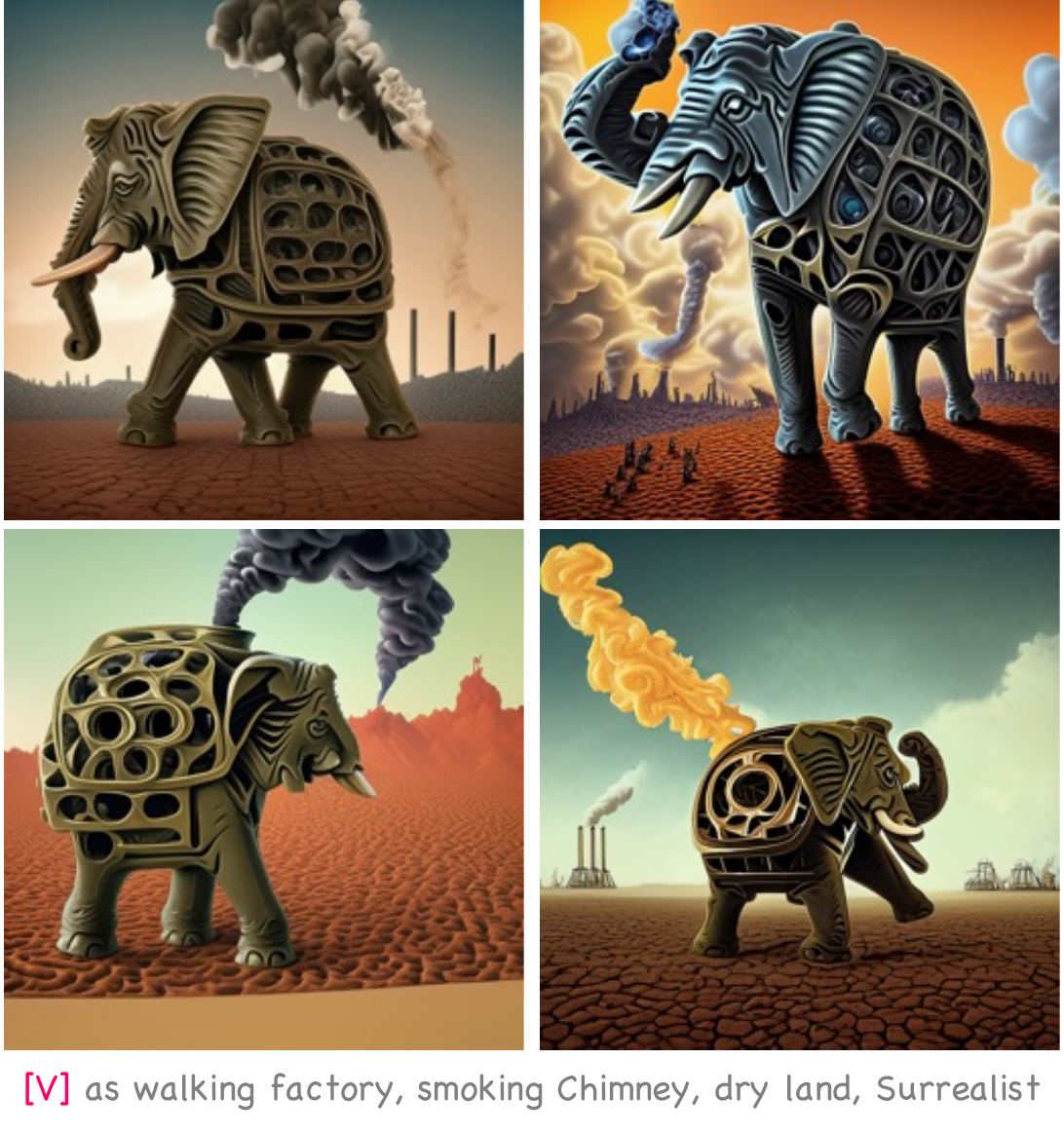} & \includegraphics[width=0.45\textwidth]{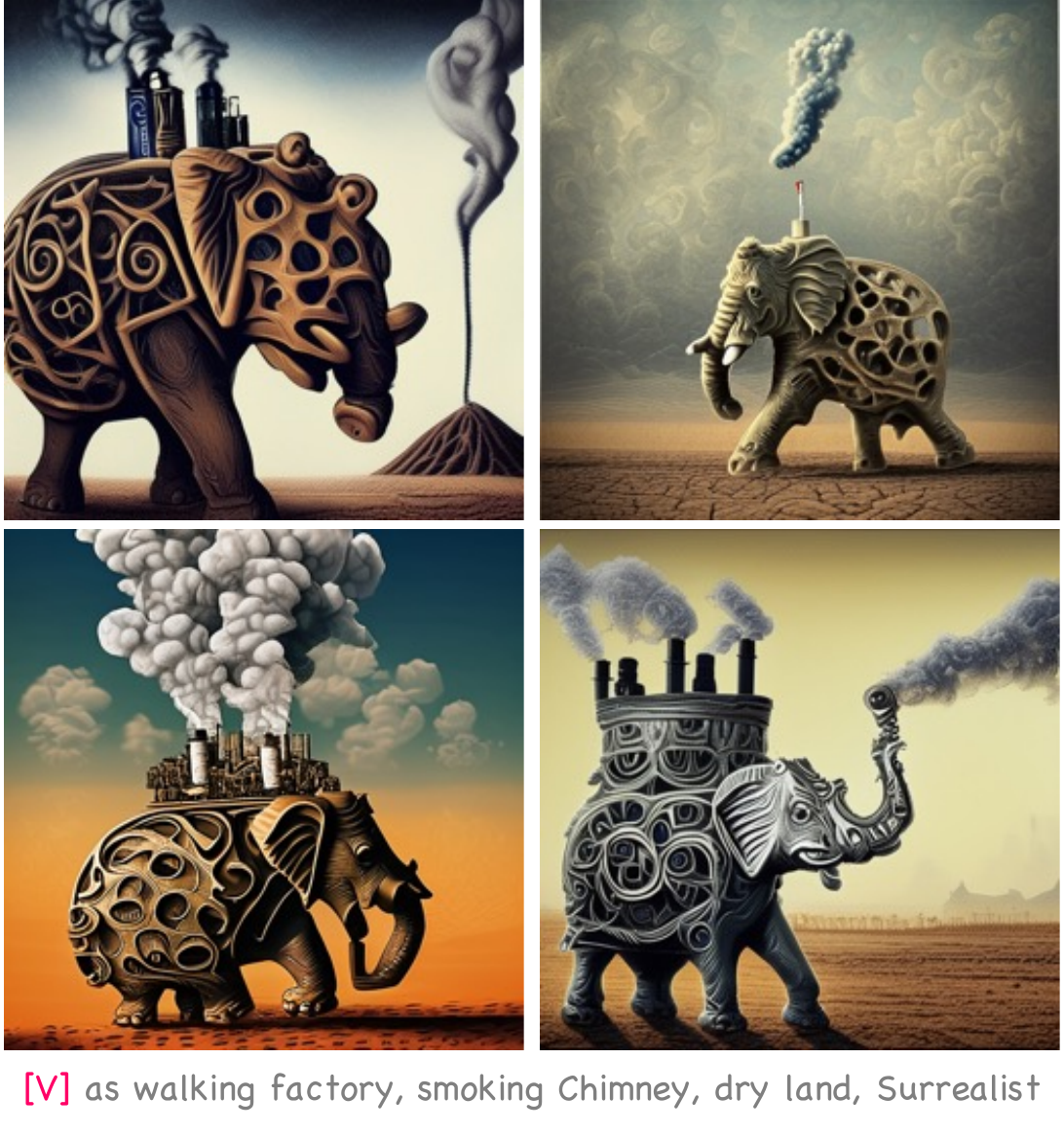}\\
         \includegraphics[width=0.45\textwidth]{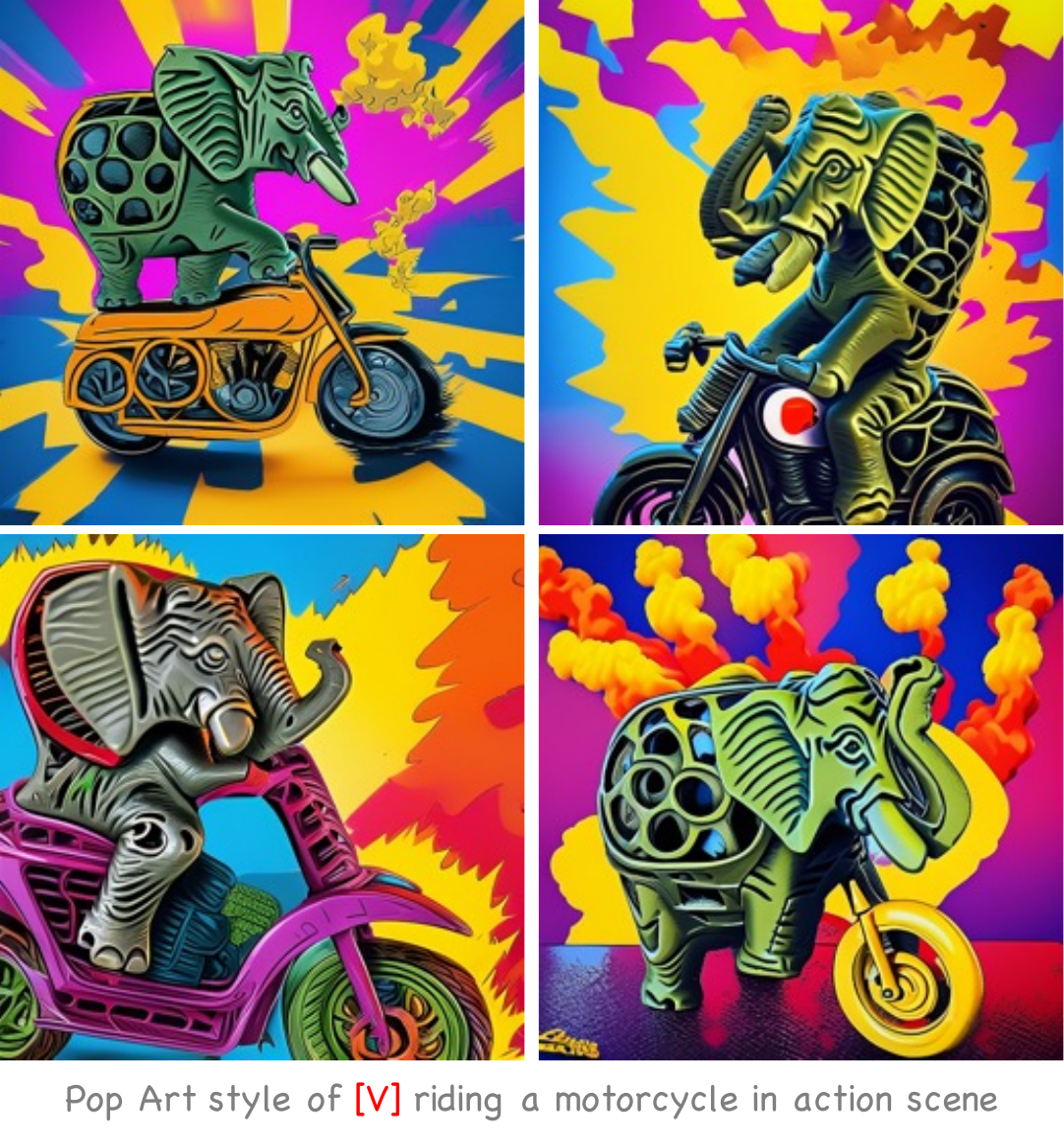}& \includegraphics[width=0.45\textwidth]{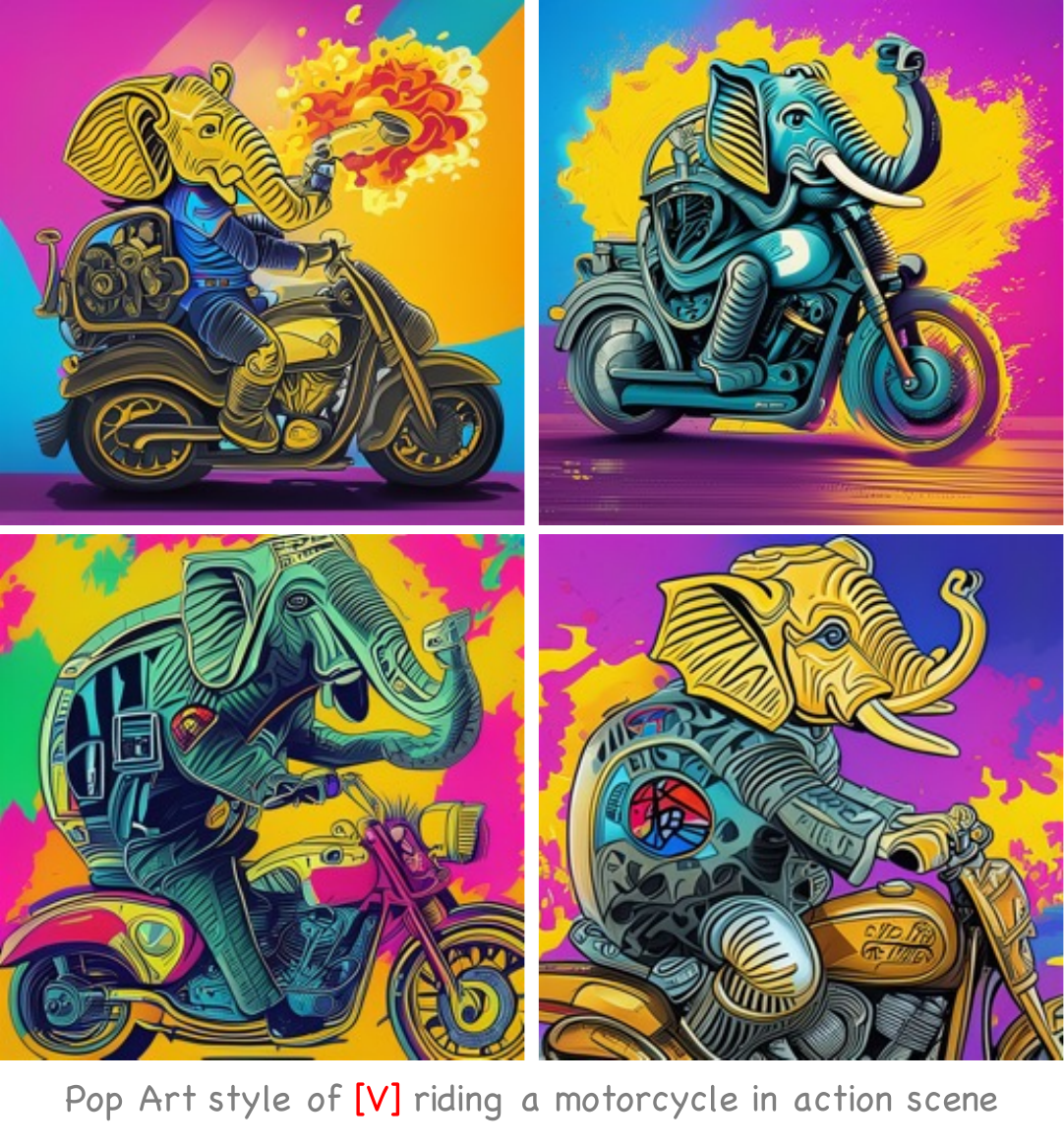} \\
         \multicolumn{2}{c}{\includegraphics[width=0.95\textwidth]{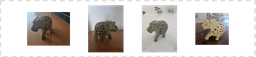}}
    \end{tabular}
    \caption{Additional Results on Larger Text-to-Image Models. Reference images appear at the bottom.}
    \label{tab:LARGE_2}
\end{table*}

\begingroup

\setlength{\tabcolsep}{2pt} %
\renewcommand{\arraystretch}{1.0} %

\begin{table*}
        \centering
        \begin{tabular}{cccccccc}
             Input &  \multicolumn{2}{c}{Ours} &  \hspace{2pt} &  TI+DB~\cite{TI, DB} &  CD~\cite{CustomDiffusion} & $P+$~\cite{PPLUS} & NeTI~\cite{NeTI} \\[5 pt]
             \includegraphics[height=2.3cm, width=2.3cm]{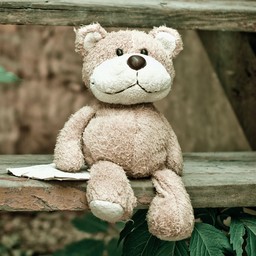} &  \includegraphics[height=2.3cm, width=2.3cm]{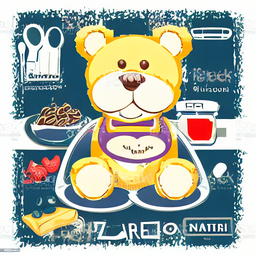} &  \includegraphics[height=2.3cm, width=2.3cm]{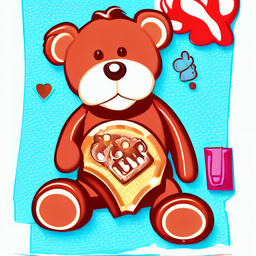} &   &  \includegraphics[height=2.3cm, width=2.3cm]{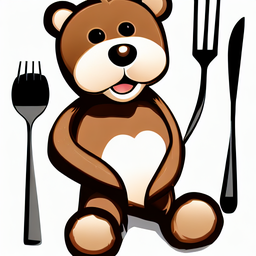} &  \includegraphics[height=2.3cm, width=2.3cm]{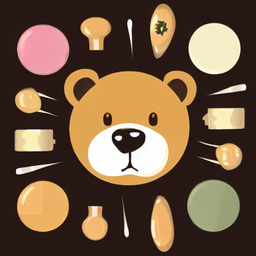} &  \includegraphics[height=2.3cm, width=2.3cm]{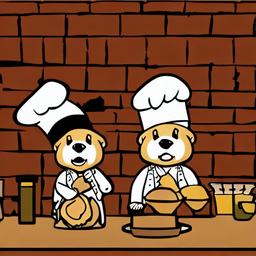} & \includegraphics[height=2.3cm, width=2.3cm]{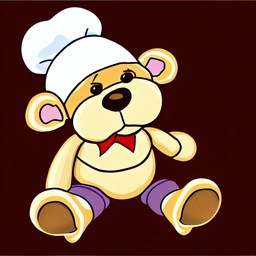} \\
             \includegraphics[height=2.3cm, width=2.3cm]{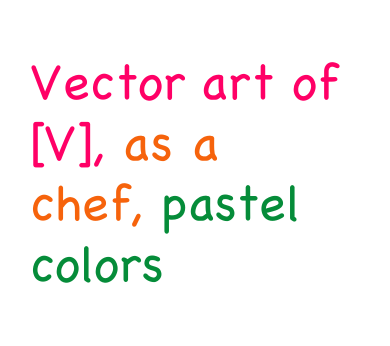} &  \includegraphics[height=2.3cm, width=2.3cm]{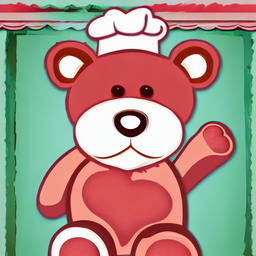} &  \includegraphics[height=2.3cm, width=2.3cm]{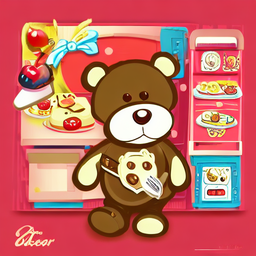} &   &  \includegraphics[height=2.3cm, width=2.3cm]{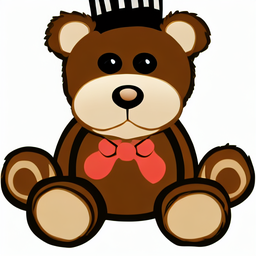} &  \includegraphics[height=2.3cm, width=2.3cm]{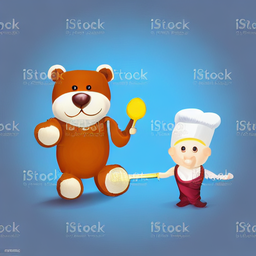} &  \includegraphics[height=2.3cm, width=2.3cm]{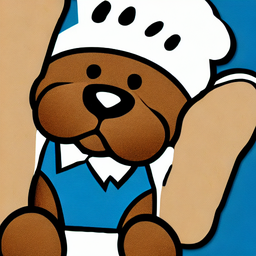} & \includegraphics[height=2.3cm, width=2.3cm]{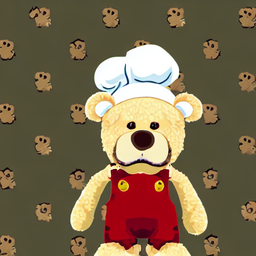} \\[4pt]

             \includegraphics[height=2.3cm, width=2.3cm]{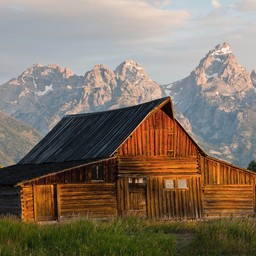} &  \includegraphics[height=2.3cm, width=2.3cm]{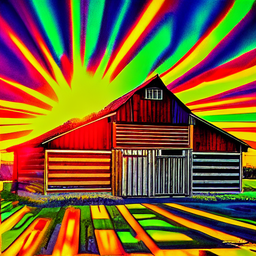} &  \includegraphics[height=2.3cm, width=2.3cm]{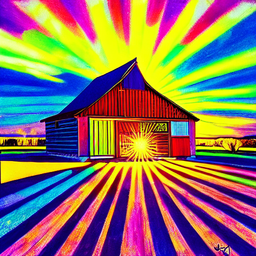} &   &  \includegraphics[height=2.3cm, width=2.3cm]{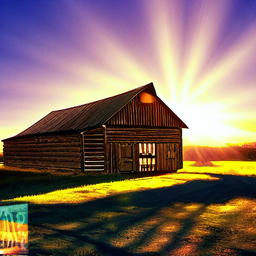} &  \includegraphics[height=2.3cm, width=2.3cm]{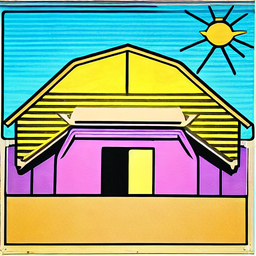} &  \includegraphics[height=2.3cm, width=2.3cm]{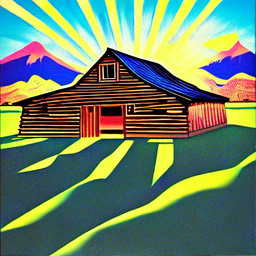} & \includegraphics[height=2.3cm, width=2.3cm]{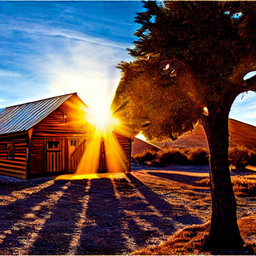} \\
             \includegraphics[height=2.3cm, width=2.3cm]{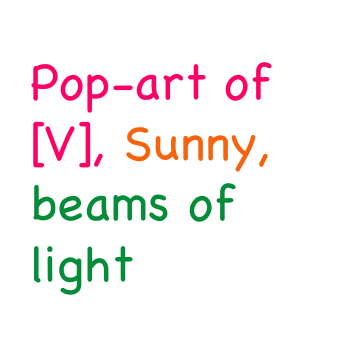} &  \includegraphics[height=2.3cm, width=2.3cm]{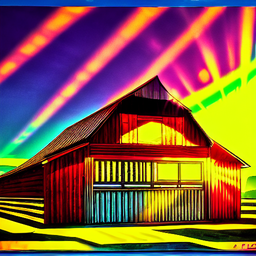} &  \includegraphics[height=2.3cm, width=2.3cm]{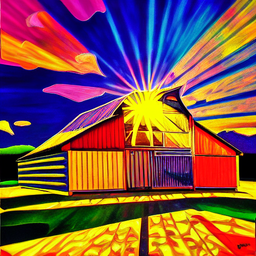} &   &  \includegraphics[height=2.3cm, width=2.3cm]{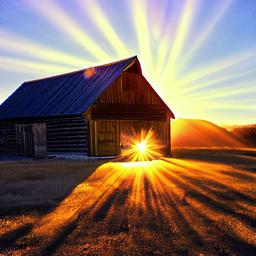} &  \includegraphics[height=2.3cm, width=2.3cm]{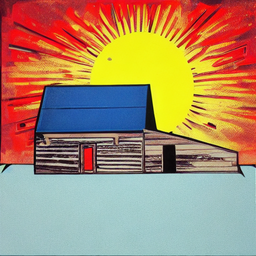} &  \includegraphics[height=2.3cm, width=2.3cm]{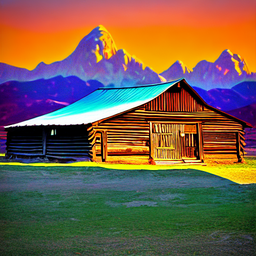} & \includegraphics[height=2.3cm, width=2.3cm]{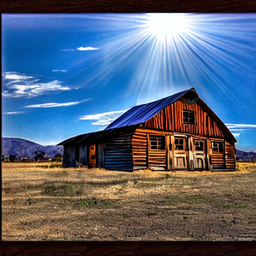} \\[4pt]

             \includegraphics[height=2.3cm, width=2.3cm]{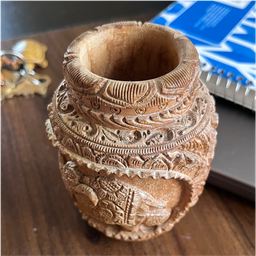} &  \includegraphics[height=2.3cm, width=2.3cm]{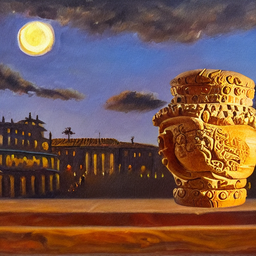} &  \includegraphics[height=2.3cm, width=2.3cm]{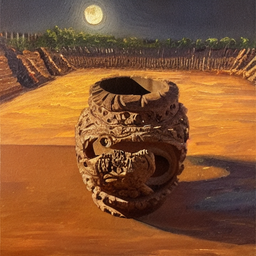} &   &  \includegraphics[height=2.3cm, width=2.3cm]{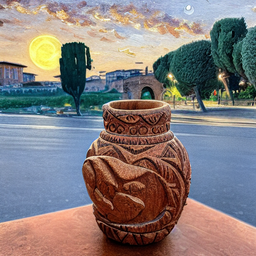} &  \includegraphics[height=2.3cm, width=2.3cm]{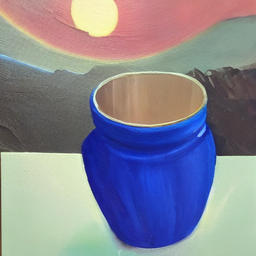} &  \includegraphics[height=2.3cm, width=2.3cm]{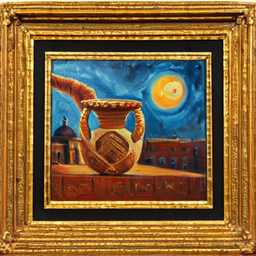} & \includegraphics[height=2.3cm, width=2.3cm]{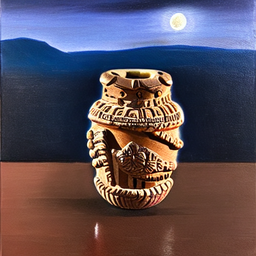} \\
             \includegraphics[height=2.3cm, width=2.3cm]{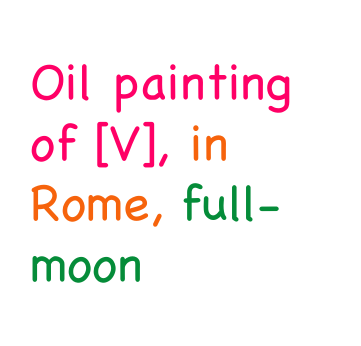} &  \includegraphics[height=2.3cm, width=2.3cm]{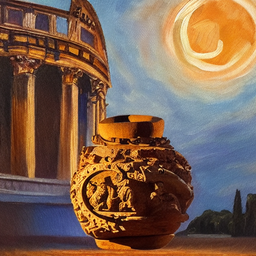} &  \includegraphics[height=2.3cm, width=2.3cm]{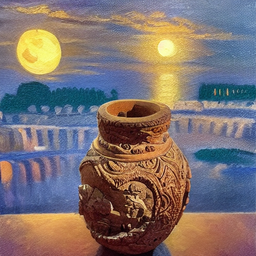} &   &  \includegraphics[height=2.3cm, width=2.3cm]{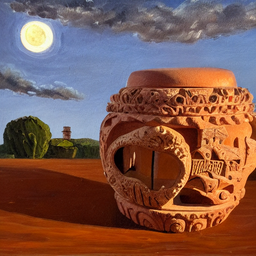} &  \includegraphics[height=2.3cm, width=2.3cm]{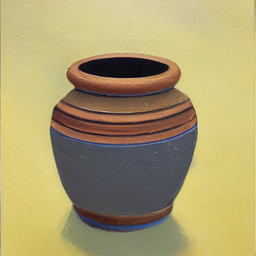} &  \includegraphics[height=2.3cm, width=2.3cm]{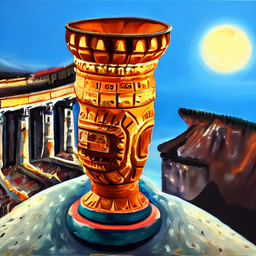} & \includegraphics[height=2.3cm, width=2.3cm]{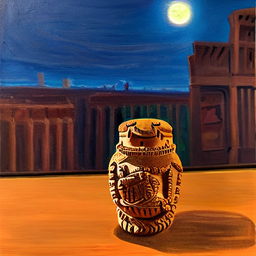} \\[4pt]

             \includegraphics[height=2.3cm, width=2.3cm]{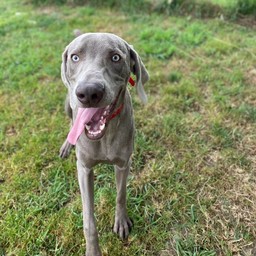} &  \includegraphics[height=2.3cm, width=2.3cm]{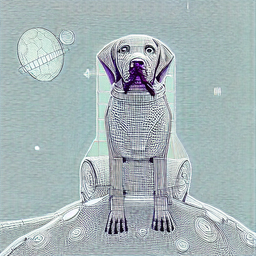} &  \includegraphics[height=2.3cm, width=2.3cm]{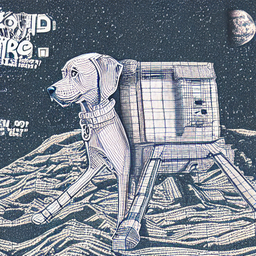} &   &  \includegraphics[height=2.3cm, width=2.3cm]{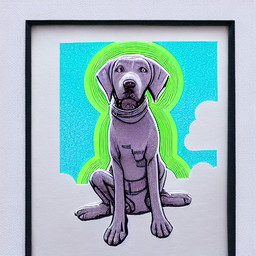} &  \includegraphics[height=2.3cm, width=2.3cm]{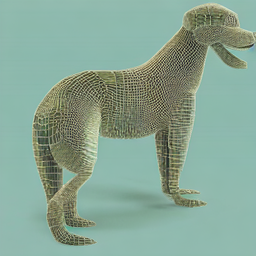} &  \includegraphics[height=2.3cm, width=2.3cm]{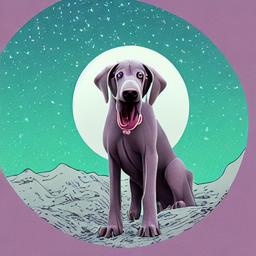} & \includegraphics[height=2.3cm, width=2.3cm]{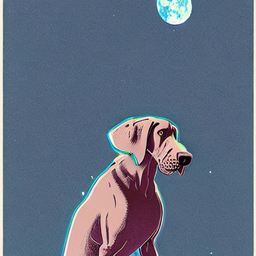} \\
             \includegraphics[height=2.3cm, width=2.3cm]{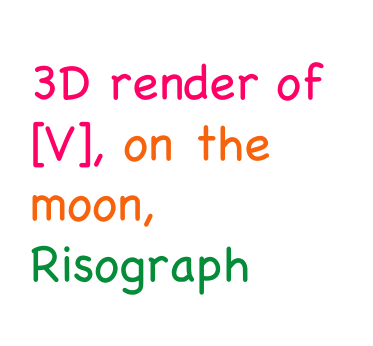} &  \includegraphics[height=2.3cm, width=2.3cm]{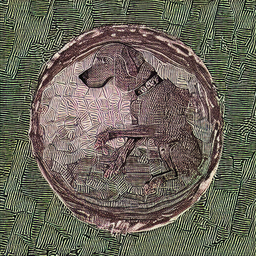} &  \includegraphics[height=2.3cm, width=2.3cm]{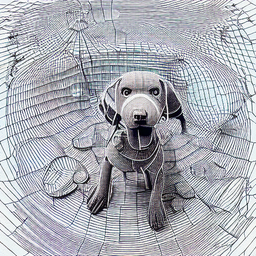} &   &  \includegraphics[height=2.3cm, width=2.3cm]{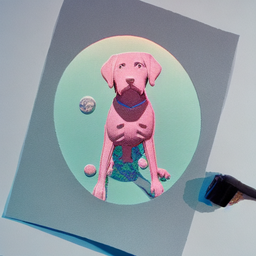} &  \includegraphics[height=2.3cm, width=2.3cm]{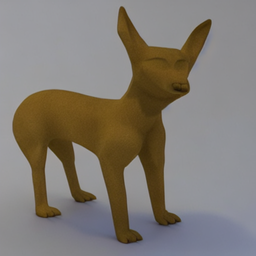} &  \includegraphics[height=2.3cm, width=2.3cm]{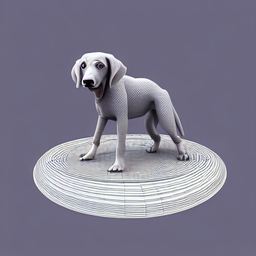} & \includegraphics[height=2.3cm, width=2.3cm]{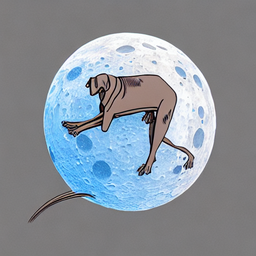} \\[4pt]

        \end{tabular}
\caption{Additional qualitative comparison in multi-shot setting.}
\label{fig:multishot_qualitative_additional}
\end{table*}

\endgroup

\begingroup

\setlength{\tabcolsep}{2pt} %
\renewcommand{\arraystretch}{1.0} %

\begin{table*}
        \centering
        \begin{tabular}{cccccccc}
             Input & Prompt & Ours &  \hspace{2pt} &  E4T~\cite{E4T} &  ProFusion~\cite{ProFusion} & IP-Adapter~\cite{IPAdapter} & Face0~\cite{Face0} \\[5 pt]
             \includegraphics[height=2.3cm, width=2.3cm]{figures/qualitative_comparison_faces/girl_input_img.png} & \includegraphics[height=2.3cm, width=2.3cm]{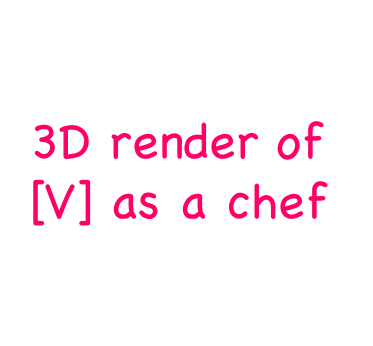} &  \includegraphics[height=2.3cm, width=2.3cm]{figures/qualitative_comparison_faces/girl_ours/girl_3d.png} &   &  \includegraphics[height=2.3cm, width=2.3cm]{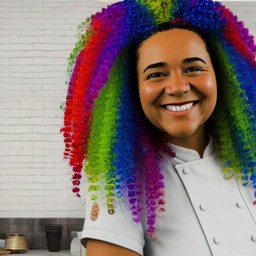} &  \includegraphics[height=2.3cm, width=2.3cm]{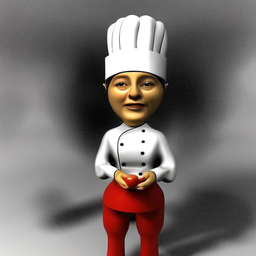} &  \includegraphics[height=2.3cm, width=2.3cm]{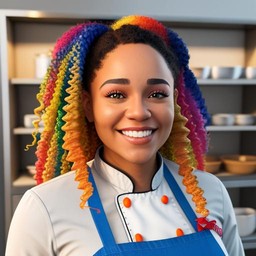} & \includegraphics[height=2.3cm, width=2.3cm]{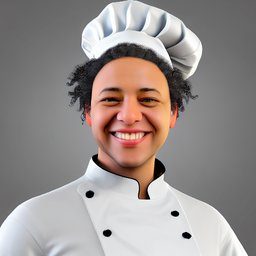} \\
             &  \includegraphics[height=2.3cm, width=2.3cm]{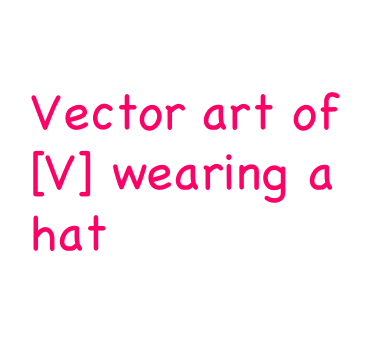} &  \includegraphics[height=2.3cm, width=2.3cm]{figures/qualitative_comparison_faces/girl_ours/girl_vectorart.png} &   &  \includegraphics[height=2.3cm, width=2.3cm]{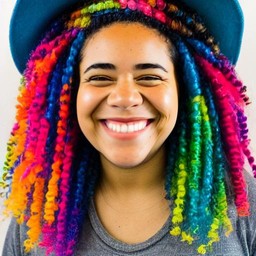} &  \includegraphics[height=2.3cm, width=2.3cm]{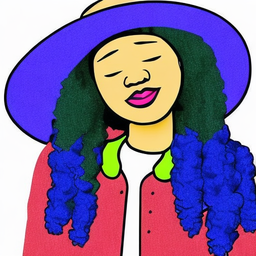} &  \includegraphics[height=2.3cm, width=2.3cm]{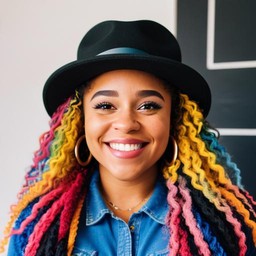} & \includegraphics[height=2.3cm, width=2.3cm]{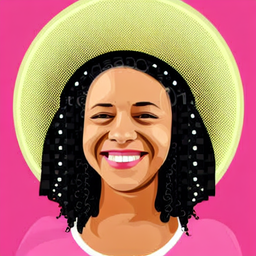} \\
             &  \includegraphics[height=2.3cm, width=2.3cm]{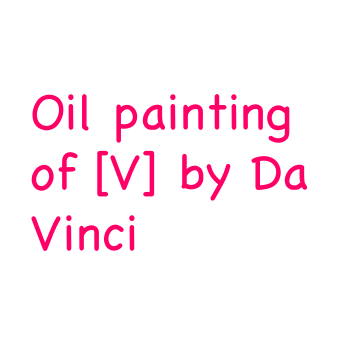} &  \includegraphics[height=2.3cm, width=2.3cm]{figures/qualitative_comparison_faces/girl_ours/girl_davinci.png} &   &  \includegraphics[height=2.3cm, width=2.3cm]{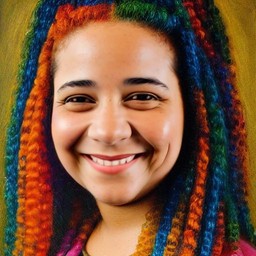} &  \includegraphics[height=2.3cm, width=2.3cm]{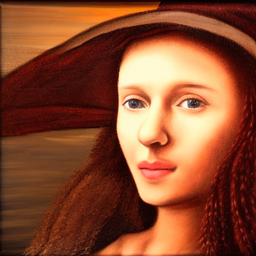} &  \includegraphics[height=2.3cm, width=2.3cm]{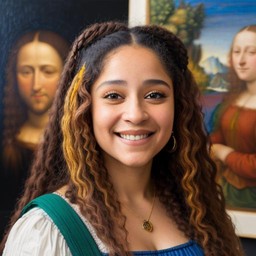} & \includegraphics[height=2.3cm, width=2.3cm]{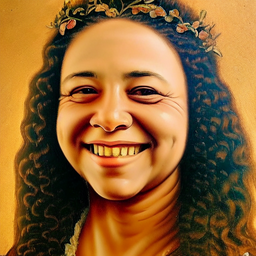} \\
             &  \includegraphics[height=2.3cm, width=2.3cm]{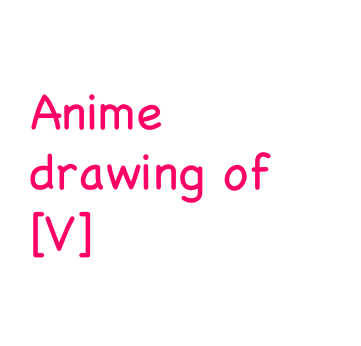} &  \includegraphics[height=2.3cm, width=2.3cm]{figures/qualitative_comparison_faces/girl_ours/girl_anime.png} &   &  \includegraphics[height=2.3cm, width=2.3cm]{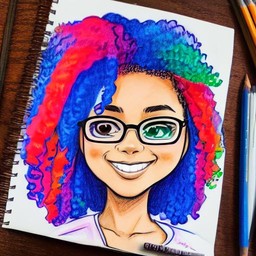} &  \includegraphics[height=2.3cm, width=2.3cm]{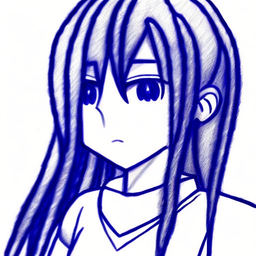} &  \includegraphics[height=2.3cm, width=2.3cm]{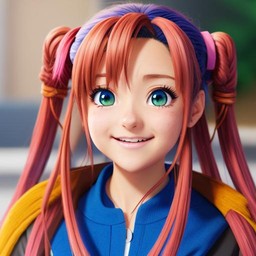} & \includegraphics[height=2.3cm, width=2.3cm]{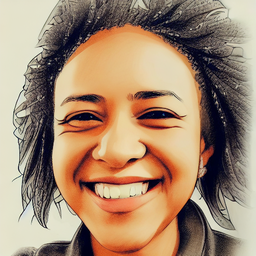} \\
             &  \includegraphics[height=2.3cm, width=2.3cm]{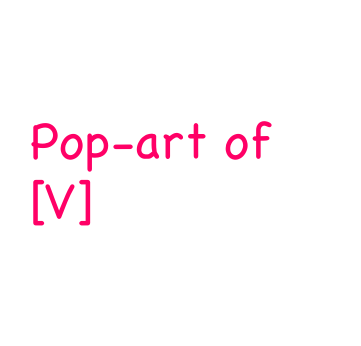} &  \includegraphics[height=2.3cm, width=2.3cm]{figures/qualitative_comparison_faces/girl_ours/girl_popart.png} &   &  \includegraphics[height=2.3cm, width=2.3cm]{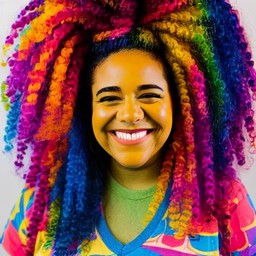} &  \includegraphics[height=2.3cm, width=2.3cm]{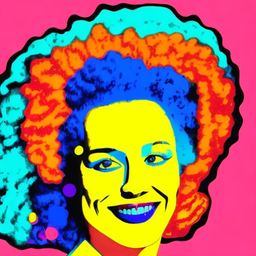} &  \includegraphics[height=2.3cm, width=2.3cm]{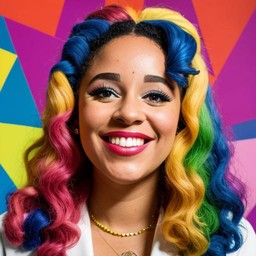} & \includegraphics[height=2.3cm, width=2.3cm]{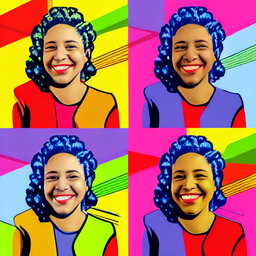} \\
             &  \includegraphics[height=2.3cm, width=2.3cm]{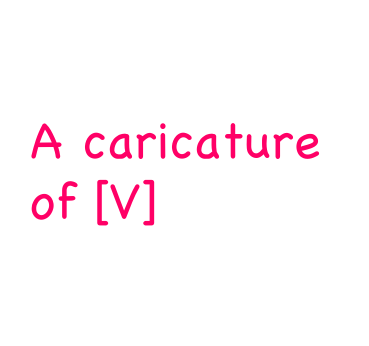} &  \includegraphics[height=2.3cm, width=2.3cm]{figures/qualitative_comparison_faces/girl_ours/girl_caricature.png} &   &  \includegraphics[height=2.3cm, width=2.3cm]{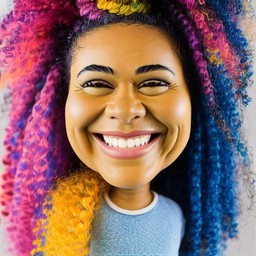} &  \includegraphics[height=2.3cm, width=2.3cm]{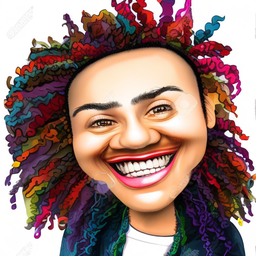} &  \includegraphics[height=2.3cm, width=2.3cm]{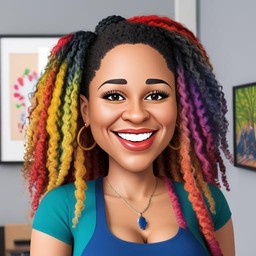} & \includegraphics[height=2.3cm, width=2.3cm]{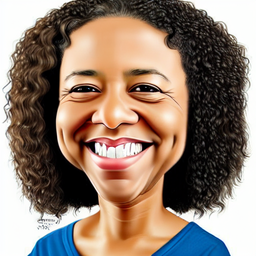} \\
              
        \end{tabular}
\caption{Single-shot setting - full qualitative comparison.}
\label{fig:sup_faces_full_girl}
\end{table*}

\endgroup

\begingroup

\setlength{\tabcolsep}{2pt} %
\renewcommand{\arraystretch}{1.0} %

\begin{table*}
        \centering
        \begin{tabular}{cccccccc}
             Input & Prompt & Ours &  \hspace{2pt} &  E4T~\cite{E4T} &  ProFusion~\cite{ProFusion} & IP-Adapter~\cite{IPAdapter} & Face0~\cite{Face0} \\[5 pt]
             \includegraphics[height=2.3cm, width=2.3cm]{figures/qualitative_comparison_faces/me_arxiv.jpg} & \includegraphics[height=2.3cm, width=2.3cm]{sup_figures/faces_full/3d_text.pdf} &  \includegraphics[height=2.3cm, width=2.3cm]{figures/qualitative_comparison_faces/guy_ours/guy_3d.png} &   &  \includegraphics[height=2.3cm, width=2.3cm]{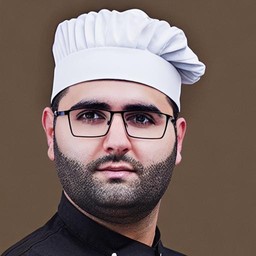} &  \includegraphics[height=2.3cm, width=2.3cm]{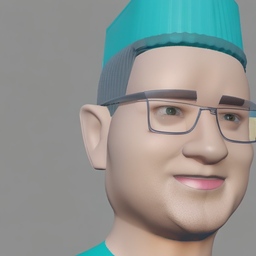} &  \includegraphics[height=2.3cm, width=2.3cm]{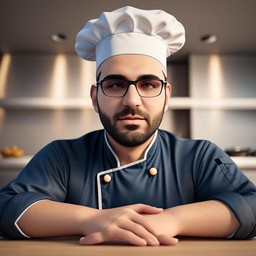} & \includegraphics[height=2.3cm, width=2.3cm]{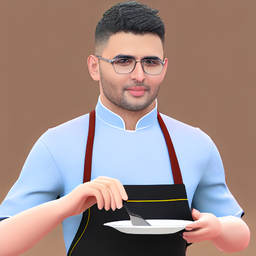} \\
             &  \includegraphics[height=2.3cm, width=2.3cm]{sup_figures/faces_full/vectorart_text.pdf} &  \includegraphics[height=2.3cm, width=2.3cm]{figures/qualitative_comparison_faces/guy_ours/guy_vectorart.png} &   &  \includegraphics[height=2.3cm, width=2.3cm]{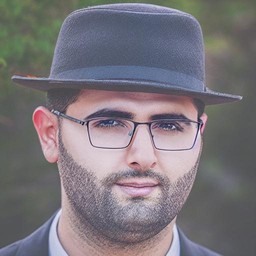} &  \includegraphics[height=2.3cm, width=2.3cm]{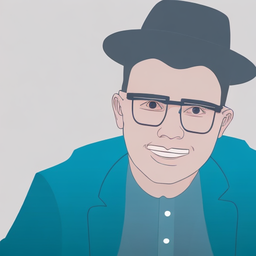} &  \includegraphics[height=2.3cm, width=2.3cm]{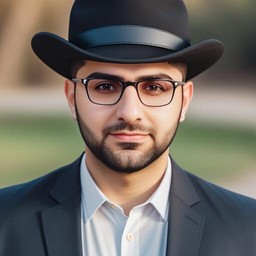} & \includegraphics[height=2.3cm, width=2.3cm]{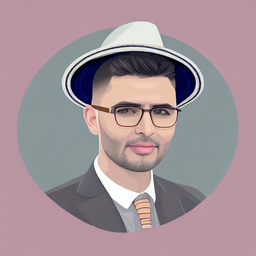} \\
             &  \includegraphics[height=2.3cm, width=2.3cm]{sup_figures/faces_full/davinci_text.pdf} &  \includegraphics[height=2.3cm, width=2.3cm]{figures/qualitative_comparison_faces/guy_ours/guy_davinci.png} &   &  \includegraphics[height=2.3cm, width=2.3cm]{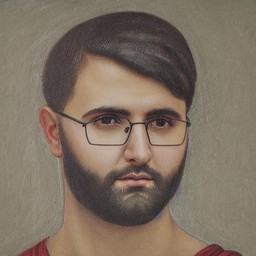} &  \includegraphics[height=2.3cm, width=2.3cm]{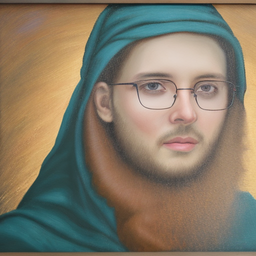} &  \includegraphics[height=2.3cm, width=2.3cm]{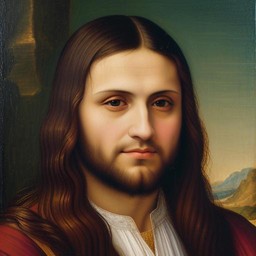} & \includegraphics[height=2.3cm, width=2.3cm]{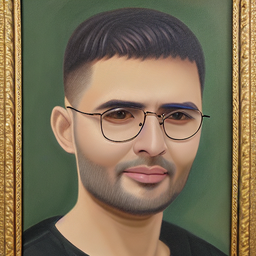} \\
             &  \includegraphics[height=2.3cm, width=2.3cm]{sup_figures/faces_full/anime_text.pdf} &  \includegraphics[height=2.3cm, width=2.3cm]{figures/qualitative_comparison_faces/guy_ours/guy_anime.png} &   &  \includegraphics[height=2.3cm, width=2.3cm]{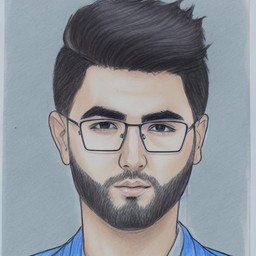} &  \includegraphics[height=2.3cm, width=2.3cm]{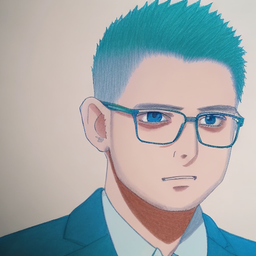} &  \includegraphics[height=2.3cm, width=2.3cm]{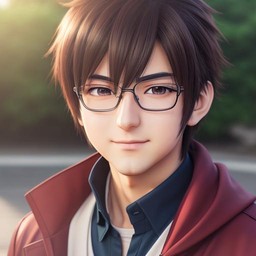} & \includegraphics[height=2.3cm, width=2.3cm]{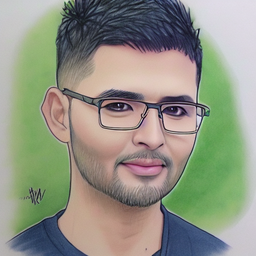} \\
             &  \includegraphics[height=2.3cm, width=2.3cm]{sup_figures/faces_full/popart_text.pdf} &  \includegraphics[height=2.3cm, width=2.3cm]{figures/qualitative_comparison_faces/guy_ours/guy_popart.png} &   &  \includegraphics[height=2.3cm, width=2.3cm]{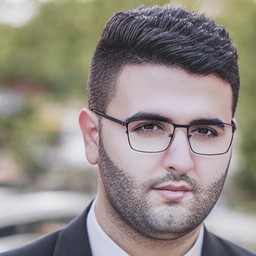} &  \includegraphics[height=2.3cm, width=2.3cm]{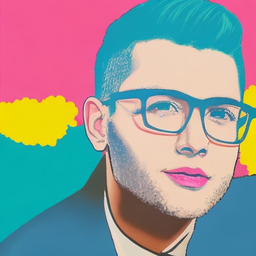} &  \includegraphics[height=2.3cm, width=2.3cm]{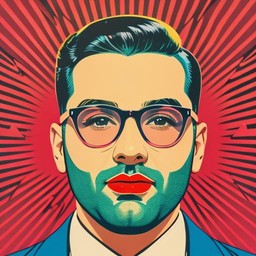} & \includegraphics[height=2.3cm, width=2.3cm]{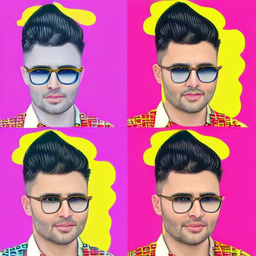} \\
             &  \includegraphics[height=2.3cm, width=2.3cm]{sup_figures/faces_full/caricature_text.pdf} &  \includegraphics[height=2.3cm, width=2.3cm]{figures/qualitative_comparison_faces/guy_ours/guy_caricature.png} &   &  \includegraphics[height=2.3cm, width=2.3cm]{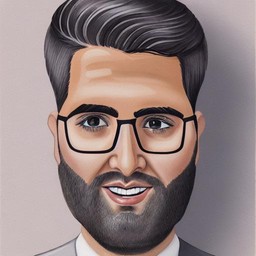} &  \includegraphics[height=2.3cm, width=2.3cm]{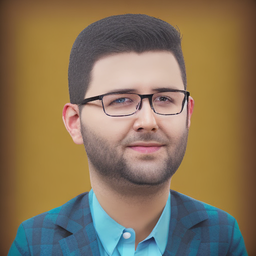} &  \includegraphics[height=2.3cm, width=2.3cm]{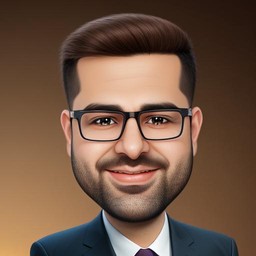} & \includegraphics[height=2.3cm, width=2.3cm]{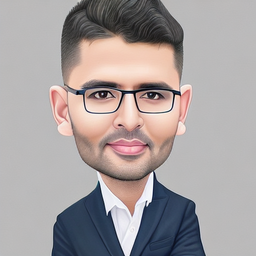} \\
              
        \end{tabular}
\caption{Single-shot setting - full qualitative comparison.}
\label{fig:sup_faces_full_guy}
\end{table*}

\endgroup

\end{document}